\documentclass[10pt,journal,compsoc]{IEEEtran}

\usepackage{times}
\usepackage{epsfig}
\usepackage{graphicx}
\usepackage{amsmath}
\usepackage{amssymb}
\usepackage{mathtools}
\usepackage{verbatim}
\usepackage{cite}
\usepackage{subfigure}
\usepackage{booktabs} % 
\usepackage{multirow}
\usepackage{array}
\usepackage{footnote}
\usepackage{ragged2e}
\newcommand{\myparagraph}[1]{\vspace{0.1em}\noindent\textbf{#1}}
\usepackage[linesnumbered,lined,vlined,ruled,commentsnumbered]{algorithm2e}
\usepackage[pagebackref=true,breaklinks=true,colorlinks,bookmarks=false]{hyperref}

\ifCLASSINFOpdf
\else
\fi
\begin{document}
\title{Meta-Transfer Learning through Hard Tasks}
% through Hard Tasks \\for Efficient Few-shot Learning}

\author{Qianru~Sun*,
        Yaoyao~Liu*,
        Zhaozheng~Chen,
        Tat-Seng Chua,
        and~Bernt~Schiele,~\IEEEmembership{Fellow,~IEEE}% <-this % stops a space
\IEEEcompsocitemizethanks{
\IEEEcompsocthanksitem Qianru Sun is the corresponding author. qianrusun@smu.edu.sg.
She and Zhaozheng Chen are with the School of Information Systems, Singapore Management University. 
Yaoyao Liu is with the School of Electrical and Information Engineering, Tianjin University. This work was done during his internship supervised by Qianru Sun.
Tat-Seng Chua is with the School of Computing, National University of Singapore.
Bernt Schiele is with the Department of Computer Vision and Machine Learning, Max-Plank Institute for Informatics.
\IEEEcompsocthanksitem * indicates equal contribution.
}
\thanks{Manuscript received on Sep 29th, 2019.}
}


\markboth{Under review}%
{Sun \MakeLowercase{\textit{et al.}}: 
Meta-Transfer Learning through Hard Tasks}



\IEEEtitleabstractindextext{%

\begin{abstract}
\justifying
Meta-learning has been proposed as a framework to address the challenging few-shot learning setting. The key idea is to leverage a large number of similar few-shot tasks in order to learn how to adapt a base-learner to a new task for which only a few labeled samples are available. As deep neural networks (DNNs) tend to overfit using a few samples only, typical meta-learning models use shallow neural networks, thus limiting its effectiveness.
In order to achieve top performance, some recent works tried to use the DNNs pre-trained on large-scale datasets but mostly in straight-forward manners, e.g., (1)~taking their weights as a warm start of meta-training, and (2)~freezing their convolutional layers as the feature extractor of base-learners.
In this paper, we propose a novel approach called \textbf{meta-transfer learning (MTL)} which \textbf{learns to transfer the weights of a deep NN} for few-shot learning tasks.
Specifically, \emph{meta} refers to training multiple tasks, and \emph{transfer} is achieved by learning scaling and shifting functions of DNN weights for each task.
In addition, we introduce the \textbf{hard task (HT) meta-batch} scheme as an effective learning curriculum that further boosts the learning efficiency of MTL.
We conduct few-shot learning experiments and report top performance for five-class few-shot recognition tasks on three challenging benchmarks: miniImageNet, tieredImageNet and Fewshot-CIFAR100 (FC100). 
Extensive comparisons to related works validate that our MTL approach trained with the proposed HT meta-batch scheme achieves top performance. An ablation study also shows that both components contribute to fast convergence and high accuracy.

Compared to the conference version of the paper~\cite{SunCVPR2019}, this version additionally presents
(1)~the analysis of using different DNN architectures, e.g., ResNet-12, ResNet-18 and ResNet-25;
(2)~the new MTL variants that adapt our scaling and shifting functions to the classical supervised and the state-of-the-art semi-supervised meta-learners, and achieve performance improvements consistently; 
(3)~the discussion of new related works since the conference version;
and (4)~the results on the larger and more challenging benchmark -- tieredImageNet~\cite{RenICLR2018_semisupervised}.

\end{abstract}

% % Note that keywords are not normally used for peerreview papers.
% \begin{IEEEkeywords}
% Meta-learning, few-shot learning, transfer learning, curriculum learning.
% \end{IEEEkeywords}
}

\maketitle

\IEEEdisplaynontitleabstractindextext

\IEEEpeerreviewmaketitle

\ifCLASSOPTIONcompsoc
\IEEEraisesectionheading{\section{Introduction}\label{sec:introduction}}
\else
\section{Introduction}
\label{sec_introduction}
\fi

%%%%app
\IEEEPARstart{A}{lthough} deep learning systems have achieved great performance when sufficient amounts of labeled data are available~\cite{Lecun2015, HeZRS16, ShelhamerLD17}, there has been growing interest in reducing the required amount of data. Few-shot learning tasks have been defined for this purpose. 
The aim is to learn new concepts from a handful of training examples, e.g. from 1 or 5 training images~\cite{FeiFeiFP06, FinnAL17, SunCVPR2019}. 
Humans tend to be highly effective in this context, often grasping the essential connection between new concepts and their own knowledge and experience, but it remains challenging for machine learning models.
For instance on the CIFAR-100 dataset, a classification model trained in the fully supervised mode achieves $76\%$ accuracy for the 100-class setting~\cite{ClevertUH15}, while the best-performing 1-shot model achieves only $45\%$ in average for the much simpler 5-class setting~\cite{SunCVPR2019}.
On the other hand, in many real-world applications we are lacking large-scale training data, as e.g. in the medical domain.
It is thus desirable to improve machine learning models in order to handle few-shot settings.

%%%%%tisser figure

\begin{figure*}[t]
  \centering
  \includegraphics[width=0.97\linewidth]{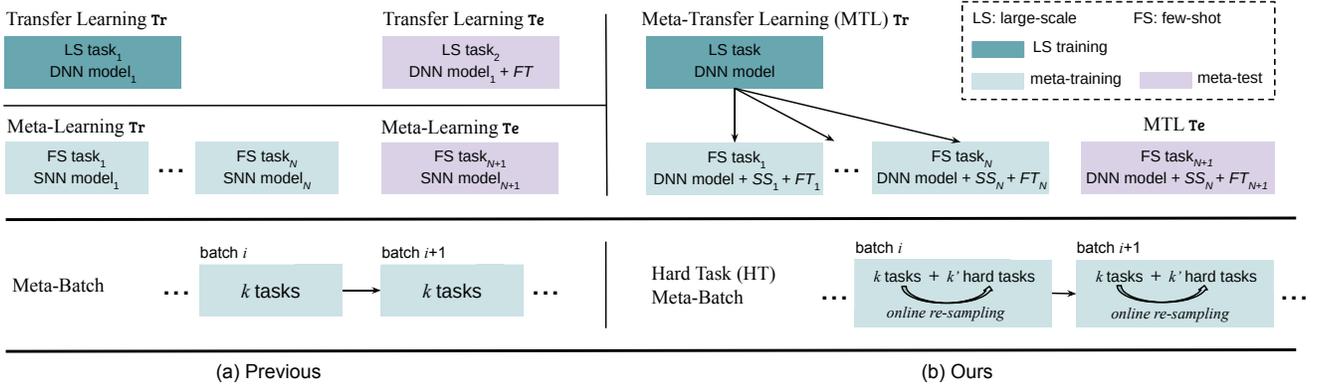}
     \caption{Meta-transfer learning (MTL) is our meta-learning paradigm and hard task (HT) meta-batch is our training strategy. The upper blocks show the differences between MTL and related methods -- transfer-learning~\cite{PanTKY11} and meta-learning~\cite{FinnAL17}.
     The bottom blocks compare the proposed HT meta-batch with the conventional meta-batch~\cite{FinnAL17}. Note that here the \emph{FT} stands for fine-tuning a classifier. \emph{SS} represents the \emph{Scaling} and \emph{Shifting} operations in our MTL method.}
  \label{main_framework_tisser}
\end{figure*}

%%%%%%%%%%%%%%%existing data aug methods
Basically, the nature of few-shot learning with very scarce training data makes it difficult to 
train powerful machine learning models for new concepts.
People explore a variety of methods in order to overcome this. 
A straight forward idea is to increase the amount of available data by data augmentation techniques~\cite{KhorevaBIBS17}.
Several methods proposed to learn a data generator e.g. conditioned on Gaussian noises~\cite{Mehrotra2017, SchwartzNIPS18, WangCVPR2018} or object attributes~\cite{XianCVPR2019a}. 
However, this data generator often under-performs when trained on few-shot data, which has been investigated by~\cite{BartunovV18}.
An alternative is to merge data from multiple tasks which, however, is often ineffective due to high variances of the data across tasks~\cite{WangCVPR2018}.

%%%%%%%%%%existing meta-learning methods
In contrast to data augmentation methods, meta-learning is a task-level learning method~\cite{Bengio92, Naik92, Thrun1998}.
It aims to transfer experience from similar few-shot learning tasks~\cite{FinnAL17, FinnNIPS2018, GrantICLR2018, FranceschiICML18, LeeICML18, ZhangNIPS2018MetaGAN, SunCVPR2019, AntoniouICLR19}. Related methods follow a unified training process that contains two loops. The inner-loop learns a base-learner for an individual task, and the outer-loop then uses the validation performance of the learned base-learner to optimize the meta-learner. 
A state-of-the-art representative method named Model-Agnostic Meta-Learning (MAML) learns to search for the optimal initialization state to fast adapt a base-learner to a new task~\cite{FinnAL17}.
Its task-agnostic property makes it possible to generalize to few-shot supervised/semi-supervised learning as well as unsupervised reinforcement learning~\cite{FinnAL17, GrantICLR2018, FinnNIPS2018, AntoniouICLR19, ZhangNIPS2018MetaGAN, RusuICLR2019}.
However, in our view, there are two main limitations of this type of approaches limiting their effectiveness: i)~these methods usually require a large number of similar tasks for meta-training which is costly; and ii)~each task is typically modeled by a low-complexity base-learner, such as a shallow neural network (SNN), to avoid model overfitting to few-shot training data, thus being unable to deploy deeper and more powerful network architectures. 
For example, for the miniImageNet dataset~\cite{VinyalsBLKW16}, MAML uses a \emph{shallow} CNN with only $4$ CONV layers and its optimal performance was obtained by learning on $240k$ tasks ($60k$ iterations in total and each meta-batch contains $4$ tasks).

%%%%%%% ours
In this paper, we propose a novel meta-learning method called \textbf{meta-transfer learning (MTL)} leveraging the advantages of both transfer learning and meta-learning (see conceptual comparison of related methods in the upper block of Figure~\ref{main_framework_tisser}). 
In a nutshell, MTL is a novel learning method that helps deep neural networks (DNNs) converge faster while reducing the probability to overfit when training on few labeled data only.  
In particular, ``transfer'' means that DNN weights trained on large-scale data can be used in other tasks by two light-weight neuron operations: \emph{Scaling} and \emph{Shifting} (\emph{SS}), i.e. $\alpha X+\beta$. ``Meta'' means that the parameters of these operations can be viewed as hyper-parameters trained on few-shot learning tasks~\cite{MunkhdalaiICML2017, LiICML2018, FranceschiICML18}.
First, large-scale trained DNN weights offer a good initialization, enabling fast convergence of MTL with fewer tasks, e.g., only $8k$ tasks for miniImageNet~\cite{VinyalsBLKW16}, $30$ times fewer than MAML~\cite{FinnAL17}.
Second, light-weight operations on DNN neurons have less parameters to learn, e.g., less than $\tfrac{2}{49}$ if considering neurons of size $7\times 7$ ($\tfrac{1}{49}$ for $\alpha$ and $<\tfrac{1}{49}$ for $\beta$), reducing the chance of overfitting to few-shot data.
Third, these operations keep those trained DNN weights unchanged, and thus avoid the problem of ``catastrophic forgetting'' which means forgetting general patterns when adapting to a specific task~\cite{LopezPazNIPS17, McCloskey1989}.
%%%% new added point
Finally, these operations are conducted on the convolutional layers mostly working for image feature extraction, thus can generalize well to a variety of few-shot learning models, e.g., MAML~\cite{FinnAL17}, MatchingNet~\cite{VinyalsBLKW16}, ProtoNet~\cite{SnellSZ17}, RelationNet~\cite{SungCVPR2018}, and Baseline++~\cite{chen19closerfewshot}.

The second main contribution of this paper is an effective meta-training curriculum.
Curriculum learning~\cite{BengioLCW09} and hard negative mining~\cite{ShrivastavaGG16} both suggest that faster convergence and stronger performance can be achieved by better arrangement of training data.
Inspired by these ideas, we design our \textbf{hard task (HT) meta-batch} strategy to offer a challenging but effective learning curriculum.
As shown in the bottom rows of Figure~\ref{main_framework_tisser}, a conventional meta-batch contains a number of random tasks~\cite{FinnAL17}, but our HT meta-batch online re-samples harder ones according to past failure tasks with lowest validation accuracy.

Our overall contribution is thus three-fold:
i)~we propose a novel \textbf{MTL} method that learns to transfer large-scale pre-trained DNN weights for solving few-shot learning tasks; 
ii)~we propose a novel \textbf{HT meta-batch} learning strategy that forces meta-transfer to ``grow faster and stronger through hardship''; and 
iii)~we conduct extensive experiments on three few-shot learning benchmarks, namely miniImageNet~\cite{VinyalsBLKW16}, tieredImageNet~\cite{RenICLR2018_semisupervised} and Fewshot-CIFAR100 (FC100)~\cite{OreshkinNIPS18}, and achieve the state-of-the-art performance. 
\textbf{Compared to the conference version} of the paper~\cite{SunCVPR2019}, this version additionally presents 
(1) the results of using different DNN architectures, e.g., ResNet-12, ResNet-18 and ResNet-25, %
(2) new MTL variants that combine our scaling and shifting functions with the classical supervised and the state-of-the-art semi-supervised meta-learners models to achieve top performance in respective scenarios, compared to post-conference works, 
(3) discussion of new related works since the conference version, 
and (4) results on the larger and more challenging benchmark -- tieredImageNet~\cite{RenICLR2018_semisupervised}. 
Our Tensorflow and Pytorch codes are open-sourced at \href{https://github.com/yaoyao-liu/meta-transfer-learning}{github.com/yaoyao-liu/meta-transfer-learning}.

\section{Related work}
Research literature on few-shot learning exhibits great diversity, spanning from data augmentation~\cite{Mehrotra2017, SchwartzNIPS18, WangCVPR2018, XianCVPR2019a} to supervised meta-learning~\cite{Hinton1987, Thrun1998}.
In this paper, we focus on the meta-learning based methods most relevant to ours and compared to in the experiments.
Besides, we borrow the idea of transfer learning when leveraging the large-scale pre-training step in prior to meta-transfer. For task sampling, our HT meta-batch scheme is related to curriculum learning and hard negative sampling methods.

\myparagraph{Meta-learning.}
%
% In this section, we focus on methods using the supervised meta-learning paradigm~\cite{Hinton1987, Thrun1998} most relevant to ours and compared to in the experiments. 
%
We can divide meta-learning methods into three categories.
1)~\emph{Metric learning} methods learn a similarity space in which learning is particularly efficient for few-shot training examples.
Examples of distance metrics include cosine similarity~\cite{VinyalsBLKW16, chen19closerfewshot}, Euclidean distance to the prototypical representation of a class~\cite{SnellSZ17}, CNN-based relation module~\cite{SungCVPR2018}, ridge regression based~\cite{BertinettoICLR2019ridge}, and graph model based~\cite{SatorrasICLR2018graph, LiuICLR2019transductive}. Some recent works also tried to generate task-specific feature representation for few-shot episodes based on metric learning, like~\cite{LiICML2019best1shotResult, LiCVPR2019bestResult}
2)~\emph{Memory network} methods learn to store ``experience'' when learning seen tasks and then generalize it to unseen tasks.
The key idea is to design a model specifically for fast learning with a few training steps. 
A family of model architectures use external memory storage include Neural Turing Machines~\cite{SantoroBBWL16}, Meta Networks~\cite{MunkhdalaiICML2017}, Neural Attentive Learner (SNAIL)~\cite{MishraICLR2018}, and Task Dependent Adaptive Metric (TADAM)~\cite{OreshkinNIPS18}. For test, general meta memory and specific task information are combined to make predictions in neural networks.
3)~\emph{Gradient descent} based meta-learning methods intend for adjusting the optimization algorithm so that the model can converge within a small number of optimization steps (with a few examples).
%
%
% \cite{FinnAL17, RaviICLR2017, LeeICML18, GrantICLR2018, ZhangNIPS2018MetaGAN, RusuICLR2019} 
The optimization algorithm can be explicitly modeled with two learning loops that outer-loop has a \emph{meta-learner} that learns to adapt an inner-loop \emph{base-learner} (to few-shot examples) through different tasks.
For example, Ravi \emph{et al}. ~\cite{RaviICLR2017} introduced a method that compresses the base-learners' parameter space in an LSTM meta-learner.
Rusu \emph{et al}.~\cite{RusuICLR2019} designed a classifier generator as the meta-learner which output parameters for each specific base-learning task.
Finn \emph{et al}.~\cite{FinnAL17} proposed a meta-learner called MAML that learns to effectively initialize a base-learner for a new task. 
Lee \emph{et al}.~\cite{LeeCVPR19svm} presented a meta-learning approach with convex base-learners for few-shot tasks. 
Other related works in this category include Hierarchical Bayesian model~\cite{GrantICLR2018}, Bilevel Programming~\cite{FranceschiICML18}, and GAN based meta model~\cite{ZhangNIPS2018MetaGAN}.

Among them, MAML is a fairly general optimization algorithm, compatible with any model that learns through gradient descent.
Its meta-learner optimization is done by gradient descent using the validation loss of the base-learner.
It is closely related to our MTL.
An important difference is that MTL leverages transfer learning and benefits from referencing neuron knowledge in pre-trained deep nets. Although MAML can start from 
a pre-trained network, its element-wise fine-tuning makes it hard to learn deep nets without overfitting (validated in our experiments).

\begin{figure*}[htp]
  \centering
  \includegraphics[width=0.99\linewidth]{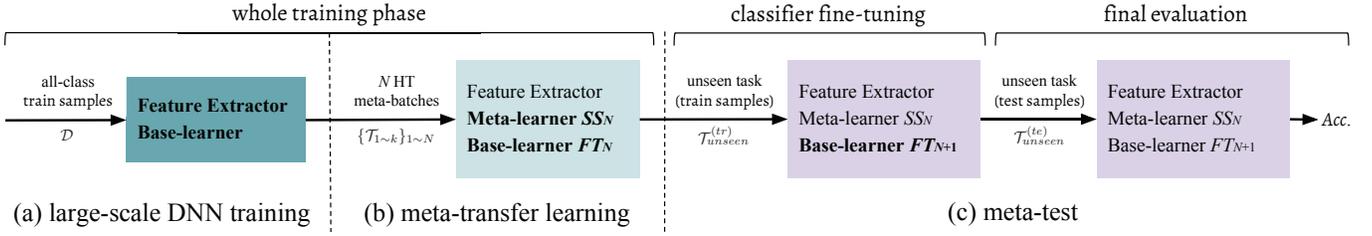}
     \caption{The pipeline of our proposed few-shot learning method, including three phases: (a) DNN training on large-scale data, i.e. using all training datapoints (Section~\ref{sec_large_scale_pretrain}); (b) Meta-transfer learning (MTL) that learns the parameters of \emph{Scaling} and \emph{Shifting} (\emph{SS}), based on the pre-trained feature extractor (Section~\ref{sec_meta_transfer}). Learning is scheduled by the proposed HT meta-batch (Section~\ref{sec_HT}); and (c) meta-test is done for an unseen task which consists of a base-learner (classifier) \emph{Fine-Tuning} (\emph{FT}) stage and a final evaluation stage, described in the last paragraph in Section~\ref{sec_preli}. 
     Input data are along with arrows. Modules with names in bold get updated at corresponding phases.}
  \label{main_framework_meta_transf_hard_task}
\end{figure*}

\myparagraph{Transfer learning}
Transfer learning or knowledge transfer has the goal to transfer the information of trained models to solve unknown tasks, thereby reducing the effort to collect new training data.
%%%
\emph{What} and \emph{how} to transfer are key issues to be addressed. Different methods are applied to different source-target domains and bridge different transfer knowledge~\cite{PanTKY11, YangICDM07, WeiICML2018, AmirCVPR18, Sun_2017_CVPR}.
For deep models, a powerful transfer method is adapting a pre-trained model for a new task, often called \emph{fine-tuning} (\emph{FT}). Models pre-trained on large-scale datasets have proven to generalize better than randomly initialized ones \cite{Erhan10}.
Another popular transfer method is taking pre-trained networks as backbone and adding high-level functions, e.g. for object detection \cite{HuangCVPR017} and image segmentation \cite{He_MaskRCNN17,ChenPAMI18}.
Besides, the knowledge to transfer can be from multi-modal category models, e.g. the word embedding models used for zero-shot learning~\cite{RohrbachNIPS13transfer, XianCVPR2019a} and trained attribute models used for social relationship recognition~\cite{Sun_2017_CVPR}.

In this paper, our meta-transfer learning leverages the idea of transferring pre-trained weights and our model meta-learns how to effectively transfer. 
The large-scale trained DNN weights are \emph{what} to transfer, and the operations of \emph{Scaling} and \emph{Shifting} indicate \emph{how} to transfer.
Some few-shot learning methods have been proposed to utilize pre-trained DNNs \cite{Keshari18, MishraICLR2018, QiaoCVPR2018, ScottNIPS2018, RusuICLR2019}. Typically, DNN weights are either fixed for feature extraction or simply fine-tuned for each task, while we learn a meta-transfer learner through all tasks, which is different in terms of the underlying learning paradigm. More importantly, our approach can generalize to existing few-shot learning models whose image features are extracted from DNNs on different architectures, for which we conduct extensive experiments in Section~\ref{sec_exp}.

\myparagraph{Curriculum learning \& Hard sample mining}
Curriculum learning was proposed by Bengio \emph{et al}. \cite{BengioLCW09} and is popular for multi-task learning~\cite{SarafianosGNK17, WeinshallCA18, GravesICML2017}.
They showed that instead of observing samples at random it is better to organize samples in a meaningful way so that fast convergence, effective learning and better generalization can be achieved.
Kumar \emph{et al}.~\cite{KumarPK10} introduced an iterative self-paced learning algorithm where each iteration simultaneously selects easy samples and learns a new parameter vector. Intuitively, the
curriculum is determined by the pupil’s abilities rather than being fixed by a teacher.
Pentina \emph{et al}. \cite{PentinaCVPR15} use adaptive SVM classifiers to evaluate task difficulty for later organization. 
Most recently, Jiang \emph{et al}.~\cite{JiangICML2018mentor} designed a MentorNet that provides a ``curriculum'', i.e., sample weighting scheme, for StudentNet to focus on the labels which are probably correct. The trained MentorNet can be directly applied for the training of StudentNet on a new dataset. 
Differently, our MTL method does task difficulty evaluation online at the phase of test in each task, without needing any auxiliary model.

Hard sample mining was proposed by Shrivastava \emph{et al}.~\cite{ShrivastavaGG16} for object detection with DNNs. It treats image proposals overlapped with ground truth (i.e. causing more confusions) as hard negative samples. Training on more confusing data enables the detection model to achieve higher robustness and better performance~\cite{CanevetF16, HarwoodGCRD17, DalalT05}. 
Inspired by this, we sample harder tasks online and make our MTL learner ``grow faster and stronger through more hardness''. In our experiments, we show that this can be generalized to different architectures with different meta-training operations, i.e. \emph{SS} and \emph{FT}, referring to Figure~\ref{fig_ht_meta_batch}.
%
%

%%%%%%%%%%%%%%%%%%%%%%%%%%%%%%%%%%%%%%%%%%%%%%%%%%%%
\section{Preliminary}
\label{sec_preli}

In this section, we briefly introduce the unified episodic formulation in meta-learning, following related works~\cite{VinyalsBLKW16, RaviICLR2017, FinnAL17, OreshkinNIPS18, RusuICLR2019}. 
Then, we introduce the task-level data denotations used at two phases, i.e., meta-train and meta-test.

\myparagraph{Meta-learning} 
has an episodic formulation which was proposed for tackling few-shot tasks first in~\cite{VinyalsBLKW16}.
It is different from traditional image classification, in three aspects: (1)~the main phases are not train and test but meta-train and meta-test, each of which includes training and  testing; (2)~the samples in meta-train and meta-test are not datapoints but episodes, and each episode is a few-shot classification task; and (3)~the objective is not classifying unseen datapoints but to fast adapt the meta-learned experience or knowledge to the learning of a new few-shot classification task.

The denotations of two phases, meta-train and meta-test, are as follows.
A meta-train example is a classification task $\mathcal{T}$ sampled from a distribution $p(\mathcal{T})$.
$\mathcal{T}$ is called episode, including a training split $\mathcal{T}^{(tr)}$ to optimize the base-learner, i.e., the classifiers in our model, and a test split $\mathcal{T}^{(te)}$ to optimize the meta-learner, i.e., the scaling and shifting parameters in our model. 
In particular, meta-train aims to learn from a number of episodes $\{\mathcal{T}\}$ sampled from $p(\mathcal{T})$.
An unseen task $\mathcal{T}_{unseen}$ in meta-test will start from that experience of the meta-learner and adapt the base-learner. The final evaluation is done by testing a set of unseen datapoints in $\mathcal{T}^{(te)}_{unseen}$.

\myparagraph{Meta-train phase.} 
This phase aims to learn a meta-learner from multiple episodes.
In each episode, meta-training has a two-stage optimization. 
Stage-1 is called base-learning, where the cross-entropy loss is used to optimize the parameters of the base-learner.
Stage-2 contains a feed-forward test on episode test datapoints. The test loss is used to optimize the parameters of the meta-learner.
Specifically, given an episode $\mathcal{T} \in p(\mathcal{T})$, the base-learner $\theta_\mathcal{T}$ is learned from episode training data $\mathcal{T}^{(tr)}$ and its corresponding loss $\mathcal{L}_{\mathcal{T}}(\theta_\mathcal{T}, \mathcal{T}^{(tr)})$. 
After optimizing this loss, the base-learner has parameters $\tilde{\theta}_\mathcal{T}$. 
Then, the meta-learner is updated using test loss $\mathcal{L}_{\mathcal{T}}(\tilde{\theta}_\mathcal{T}, \mathcal{T}^{(te)})$. 
After meta-training on all episodes, the meta-learner is optimized by test losses $\{\mathcal{L}_{\mathcal{T}}(\tilde{\theta}_\mathcal{T}, \mathcal{T}^{(te)})\}_{\mathcal{T} \in p(\mathcal{T})}$. Therefore, the number of meta-learner updates equals to the number of episodes.

\myparagraph{Meta-test phase.} 
This phase aims to test the performance of the trained meta-learner for fast adaptation to unseen task.
Given $\mathcal{T}_{unseen}$, the meta-learner $\tilde{\theta}_\mathcal{T}$ teaches the base-learner $\theta_{\mathcal{T}_{unseen}}$ to adapt to the objective of $\mathcal{T}_{unseen}$ by some means, e.g. through initialization~\cite{FinnAL17}. 
Then, the test result on $\mathcal{T}^{(te)}_{unseen}$ is used to evaluate the meta-learning approach. 
If there are multiple unseen tasks $\{\mathcal{T}_{unseen}\}$, the average result on $\{\mathcal{T}^{(te)}_{unseen}\}$ will be the final evaluation.

\section{Methodology}

As shown in Figure~\ref{main_framework_meta_transf_hard_task}, our method consists of three phases.
First, we train a DNN on large-scale data, e.g. on miniImageNet with $64$ classes and  $600$ samples per class~\cite{VinyalsBLKW16}, and then fix the low-level layers as Feature Extractor (Section~\ref{sec_large_scale_pretrain}).
Second, in the meta-transfer learning phase, MTL learns the \emph{Scaling} and \emph{Shifting} (\emph{SS}) parameters for the Feature Extractor neurons, enabling fast adaptation to few-shot tasks (Section~\ref{sec_meta_transfer}).
For improving the overall learning, we use our HT meta-batch strategy (Section~\ref{sec_HT}).
The training steps are detailed in Algorithm~\ref{alg_overall} and Algorithm~\ref{alg_Meta} in Section~\ref{sec_alg}.
Thirdly, the typical meta-test phase is performed, as introduced in Section~\ref{sec_preli}, and the details are given in Algorithm~\ref{alg_Meta_test}.

%%%
\subsection{DNN training on large-scale data}
\label{sec_large_scale_pretrain}

This phase is similar to the classic pre-training stage as, e.g., pre-training on Imagenet for object recognition~\cite{Russakovsky2015}.
Here, we do not consider data/domain adaptation from other datasets, and pre-train on readily available data of few-shot learning benchmarks, allowing for fair comparison with other few-shot learning methods.
Specifically, for a particular few-shot dataset, we merge all-class data $\mathcal{D}$ for pre-training.
For instance, for miniImageNet~\cite{VinyalsBLKW16}, there are totally $64$ classes in the training split $\mathcal{D}$ and each class contains $600$ samples, which we use to pre-train a $64$-class classifier.

We first randomly initialize a feature extractor $\Theta$ (e.g. CONV layers in ResNets~\cite{HeZRS16}) and a classifier $\theta$ (e.g. the last FC layer in ResNets~\cite{HeZRS16}), and then optimize them by gradient descent as follows,
\begin{equation}\label{eq_large_scale_update}
 [\Theta; \theta] =: [\Theta; \theta] - \alpha\nabla\mathcal{L}_{\mathcal{D}}\big([\Theta; \theta]\big),
\end{equation}
where $\mathcal{L}$ denotes the following empirical loss,
\begin{equation}\label{eq_large_scale_loss}
    \mathcal{L}_{\mathcal{D}}\big([\Theta; \theta]\big) = \frac{1}{|\mathcal{D}|}\sum_{(x,y)\in \mathcal{D}}l\big(f_{[\Theta; \theta]}(x), y\big),
\end{equation}
e.g. cross-entropy loss, and $\alpha$ denotes the learning rate. 
In this phase, the feature extractor $\Theta$ is learned. It will be frozen in the following meta-training and meta-test phases, as shown in Figure~\ref{main_framework_meta_transf_hard_task}. 
The learned classifier $\theta$ will be discarded, because subsequent few-shot tasks contain different classification objectives, e.g. $5$-class instead of $64$-class classification for miniImageNet~\cite{VinyalsBLKW16}.
%%%%%%%%%%%%%%%%%%%%%%%%%%%%%%%%%%%%%%%%%%%%%%%%%%%%%%%%%

\subsection{Meta-transfer learning (MTL)}
\label{sec_meta_transfer}

\begin{figure}[t]
  \centering
  \includegraphics[width=0.99\linewidth]{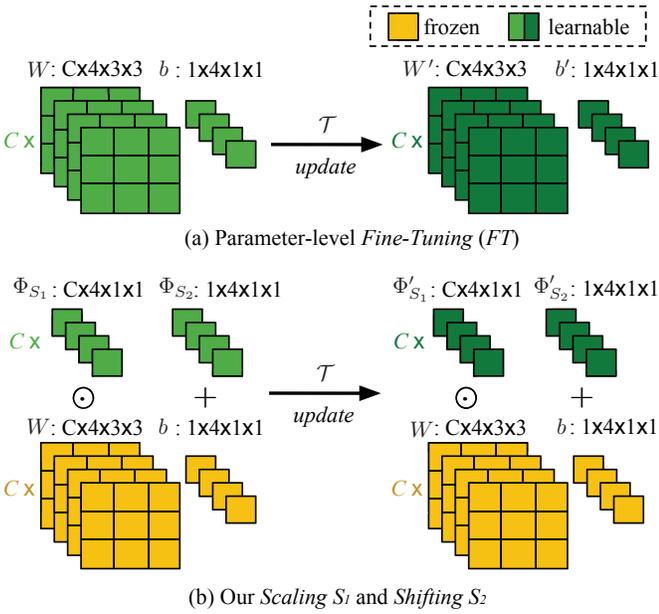}
     \caption{Two kinds of meta operations on pre-trained weights. (a) Parameter-level \emph{Fine-Tuning} (\emph{FT}) is a conventional meta-train operation used in related works such as MAML~\cite{FinnAL17}, ProtoNets~\cite{SnellSZ17} and RelationNets~\cite{SungCVPR2018}. Its update works for all neuron parameters, $W$ and $b$.
     (b) Our neuron-level \emph{Scaling} and \emph{Shifting} (\emph{SS}) operations in MTL. They reduce the number of learning parameters and avoid overfitting problems. In addition, they keep large-scale trained parameters (in yellow) frozen, preventing ``catastrophic fogetting''~\cite{LopezPazNIPS17, McCloskey1989}.
    }
  \label{main_framework_SS_FT}
  \vspace{-0.3cm}
\end{figure}

As shown in Figure~\ref{main_framework_meta_transf_hard_task}(b), our proposed meta-transfer learning (MTL) method optimizes the meta operations \emph{Scaling} and \emph{Shifting} (\emph{SS}) through HT meta-batch training (Section~\ref{sec_HT}). 
Figure~\ref{main_framework_SS_FT} visualizes the difference of updating through \emph{SS} and \emph{Fine-Tuning} (\emph{FT}).
\emph{SS} operations, denoted as $\Phi_{S_1}$ and $\Phi_{S_2}$, do not change the frozen neuron weights of $\Theta$ during learning, while \emph{FT} updates the complete $\Theta$. 

\begin{figure}[t]
  \centering
  \includegraphics[width=0.99\linewidth]{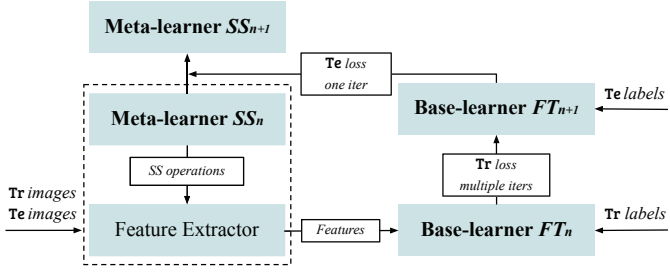}
     \caption{The computation flow on one meta-train task. It includes the update on the parameters of base-learner and meta-learner. Base-learner is initialized by the weights meta-learned from previous tasks, and its update is the conventional fine-tuning (\emph{FT}). Note that this \emph{FT} works in base-learning, thus is different with the meta-operation \emph{FT} used in the outer-loops of MAML~\cite{FinnAL17}. When this update finishes, the average loss of the test data are computed to update the MTL meta-learner, i.e. \emph{SS} parameters.}
  \label{main_framework_meta_tr}
  \vspace{-0.3cm}
\end{figure}

In the following, we expand the details of \emph{SS} operations, corresponding to Figure~\ref{main_framework_meta_transf_hard_task} (b) and Figure~\ref{main_framework_meta_tr}.
Given a task $\mathcal{T}$, the loss of $\mathcal{T}^{(tr)}$ is used to optimize the current base-learner (classifier) $\theta'$ by gradient descent:
\begin{equation}\label{eq_base_classifier}
  \theta' \gets \theta - \beta\nabla_{\theta}\mathcal{L}_{\mathcal{T}^{(tr)}}\big([\Theta; \theta], \Phi_{S_{\{1,2\}}}\big),
\end{equation}
which is different to Eq.~\ref{eq_large_scale_update}, as we do not update $\Theta$. 
Note that here $\theta$ is different to the one from the previous phase, the large-scale classifier $\theta$ in Eq.~\ref{eq_large_scale_update}. 
This $\theta$ concerns only a few classes, e.g. 5 classes, to classify each time in a novel few-shot setting. 
$\theta'$ corresponds to a temporal classifier only working in the current task, initialized by the $\theta$ optimized for the previous task (see Eq.~\ref{eq_meta_classifier}).

$\Phi_{S_1}$ is initialized by ones and $\Phi_{S_2}$ by zeros. Then, they are optimized by the test loss of $\mathcal{T}^{(te)}$ as follows,
\begin{equation}\label{eq_ss_update}
     \Phi_{S_i} =: \Phi_{S_i} - \gamma\nabla_{\Phi_{S_i}}\mathcal{L}_{\mathcal{T}^{(te)}}\big([\Theta; \theta'], \Phi_{S_{\{1,2\}}}\big), i=1, 2.
\end{equation}
In this step, $\theta$ is updated with the same learning rate $\gamma$ as in Eq.~\ref{eq_ss_update},
\begin{equation}\label{eq_meta_classifier}
  \theta =: \theta - \gamma\nabla_{\theta}\mathcal{L}_{\mathcal{T}^{(te)}}\big([\Theta; \theta'], \Phi_{S_{\{1,2\}}}\big).
\end{equation}
Re-linking to Eq.~\ref{eq_base_classifier}, we note that the above $\theta'$ comes from the last epoch of base-learning on $\mathcal{T}^{(tr)}$.

Next, we describe how we apply $\Phi_{S_{\{1,2\}}}$ to the frozen neurons as shown in Figure~\ref{main_framework_SS_FT}(b). 
Given the trained $\Theta$, for its $l$-th layer containing $K$ neurons, we have $K$ pairs of parameters, respectively as weight and bias, denoted as $\{(W_{i,k}, b_{i,k})\}$. Note that the neuron location $l, k$ will be omitted for readability.
Based on MTL, we learn $K$ pairs of scalars $\{\Phi_{S_{\{1,2\}}}\}$. 
Assuming $X$ is input, we apply $\{\Phi_{S_{\{1,2\}}}\}$ to $(W, b)$ as
\begin{equation}\label{eq_SS_operation}
    SS(X; W, b; \Phi_{S_{\{1,2\}}}) =(W\odot\Phi_{S_1}) X + (b + \Phi_{S_2}),
\end{equation}
where $\odot$ denotes the element-wise multiplication.


Taking Figure~\ref{main_framework_SS_FT}(b) as an example of a single $3\times 3$ filter, after \emph{SS} operations, this filter is scaled by $\Phi_{S_1}$ then the feature maps after convolutions are shifted by $\Phi_{S_2}$ in addition to the original bias $b$. 
Detailed steps of \emph{SS} are given in Algorithm~\ref{alg_Meta} in Section~\ref{sec_alg}.

Figure~\ref{main_framework_SS_FT}(a) shows a typical parameter-level \emph{FT} operation, which is in the meta optimization phase of our related work MAML~\cite{FinnAL17}.
It is obvious that \emph{FT} updates the complete values of $W$ and $b$, and has a large number of parameters, and our \emph{SS} reduces this number to below $\tfrac{2}{9}$ in the example of the figure. 

In summary, \emph{SS} can benefit MTL in three aspects.
1) It starts from a strong initialization based on a large-scale trained DNN, yielding fast convergence for MTL.
2) It does not change DNN weights, thereby avoiding the problem of ``catastrophic forgetting''~\cite{LopezPazNIPS17, McCloskey1989} when learning specific tasks in MTL.
3) It is light-weight, reducing the chance of overfitting of MTL in few-shot scenarios.

%%%%%%%%%%%%%%%%%%%%%%%% new
\subsection{Generalized MTL}
\label{generalized_mtl}

As introduced, key components of our meta-transfer learning include a large-scale pre-training and two meta operations -- \emph{Scaling} and \emph{Shifting} pre-trained weights.
As shown in Figure~\ref{main_framework_meta_tr}, \emph{SS} operations make the effect on feature extraction, i.e., backbone networks, for learning a new task. They are thus independent of the classifier architecture which is typically the ``head'' of the end-to-end network.

While, most metric-based and optimization-based meta learning models have same backbone ``body'' for feature extraction but different classification ``heads'' for classification.
Metric-based methods use nearest neighbor classifiers, e.g., MatchingNets~\cite{VinyalsBLKW16}, ProtoNets~\cite{SnellSZ17}, and RelationNets~\cite{SungCVPR2018}.
Optimization-based methods learn through backpropagation of unrolled gradients of multiple iterations of the classifier, and different methods have different designs for classifier, e.g., Cosine-similarity based classifier~\cite{GidarisCVPR2018, chen19closerfewshot}, Softmax classifier~\cite{FinnAL17, FinnNIPS2018, SunCVPR2019}, and SVM~\cite{LeeCVPR19svm}.

In the conference paper, we use Softmax classifier which is a simple linear FC layer in deep nets. In this journal version, we generalize meta-transfer learning method to other models.
In total, we implement five classical models including Cosine-similarity based model~\cite{GidarisCVPR2018, chen19closerfewshot}, MatchingNets~\cite{VinyalsBLKW16}, ProtoNets~\cite{SnellSZ17}, RelationNets~\cite{SungCVPR2018} and our previous Softmax model~\cite{FinnAL17, FinnNIPS2018, SunCVPR2019}.
In experiments, we show that under the same conditions of network architectures and hyperparameters, using our deep pre-trained model with \emph{SS} operations brings consistent improvements, in terms of classification accuracy as well as learning speed, over previously reported results.

%%%%%%%%%%%%%%%%%%%%%%%%%%%%%%%%%%%%%%%%%%%%%%%%%%%%%%%%%

\subsection{Hard task (HT) meta-batch}
\label{sec_HT}

\begin{figure}[t]
  \centering
  \includegraphics[width=0.99\linewidth]{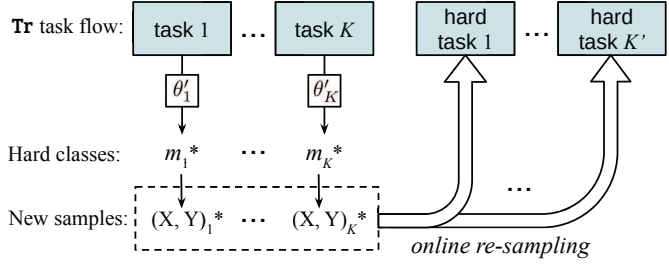}
     \caption{The computation flow of online hard task sampling. During an HT meta-batch phase, the meta-training first goes through $K$ random tasks then continues on re-sampled $K'$ hard tasks.}
  \label{figure_ht_meta_batch}
  \vspace{-0.3cm}
\end{figure}

In this section, we introduce a method to schedule hard tasks in meta-training batches. 
The conventional meta-batch is composed of randomly sampled tasks, where the randomness implies random difficulties~\cite{FinnAL17}.
In our meta-training pipeline, we intentionally pick up failure cases in each task and re-compose their data to be harder tasks for adverse re-training. The task flow is shown in Figure~\ref{figure_ht_meta_batch}.
We aim to force our meta-learner to ``grow up through hardness''. 

\myparagraph{Pipeline.}
Given a ($M$-class, $N$-shot) task $\mathcal{T}$, a meta-batch $\{\mathcal{T}_{1\sim k}\}$ contains two splits, $\mathcal{T}^{(tr)}$ and $\mathcal{T}^{(te)}$, for base-learning and test, respectively.
As shown in Algorithm~\ref{alg_Meta} line 2-5, base-learner is optimized by the loss of $\mathcal{T}^{(tr)}$ (in multiple epochs). \emph{SS} parameters are then optimized by the loss of $\mathcal{T}^{(te)}$ once.
During the loss computation on $\mathcal{T}^{(te)}$, we can also get the recognition accuracy for $M$ classes. Then, we choose the lowest accuracy $Acc_m*$ to determine the most difficult class $m*$ (also called failure class) in the current task.

After obtaining all failure classes $\{m*\}$ from $\{\mathcal{T}_{1\sim k}\}$ in current meta-batch ($k$ is the batch size), we re-sample tasks from the data indexed by $\{m*\}$. 
Specifically, we assume $p(\mathcal{T}|\{m*\})$ is the task distribution, we sample a ``harder'' task $\mathcal{T}^{hard} \in p(\mathcal{T}|\{m*\})$.
Two important details are given below.

\myparagraph{Choosing hard class $m*$.} We choose the failure class $m*$ from each task by ranking the class-level accuracies instead of fixing a threshold.
In a dynamic online setting as ours, it is more sensible to choose the hardest cases based on ranking rather than fixing a threshold ahead of time. 

\myparagraph{Two methods of hard tasking using $\{m*\}$.}
Chosen $\{m*\}$, we can re-sample tasks $\mathcal{T}^{hard}$ by (1) directly using the samples of $m*$-th class in the current task $\mathcal{T}$, or (2) indirectly using the index $m*$ to sample new samples of that class.
In fact, setting (2) considers to include more data variance of $m*$-th class and it works better than setting (1) in general. 

%%%%%%%%%%%%%%%%%%%%%%%%%%%%%%%%%%%%%%%%%%%%%%%%%%%%%%%%%
\subsection{Algorithm}
\label{sec_alg}

The training steps of our MTL approach is provided in Algorithm ~\ref{alg_overall} and Algorithm~\ref{alg_Meta}, and the meta-test steps in Algorithm~\ref{alg_Meta_test}. 

Algorithm~\ref{alg_overall} summarizes the training process of two main stages: large-scale DNN training (line 1-5) and meta-transfer learning (line 6-19). HT meta-batch re-sampling and continuous training phases are shown in line 13-18, for which the failure classes are returned by Algorithm~\ref{alg_Meta}, see line 11.

%%%%%%%%%%%%%%%%%%%%%%%%%%%%%%%%%%%%%%%%%%%%%%%%%%%%%%%%%%%

\begin{algorithm}
\caption{MTL with HT meta-batch strategy}
\label{alg_overall}
\SetAlgoLined
\SetKwInput{KwData}{Input}
\SetKwInput{KwResult}{Output}
 \KwData{Task distribution $p(\mathcal{T})$ and corresponding dataset $\mathcal{D}$, learning rates $\alpha$, $ \beta$ and $\gamma$}
 \KwResult{Feature extractor $\Theta$, base learner $\theta$, \emph{Scaling} and \emph{Shifting} parameters $\Phi_{S_{\{1,2\}}}$}
 Randomly initialize $\Theta$ and $\theta$\;
 \For{samples in $\mathcal{D}$}{
 Evaluate $\mathcal{L}_{\mathcal{D}}([\Theta; \theta])$ by Eq.~\ref{eq_large_scale_loss}\;
 Optimize $\Theta$ and $\theta$ by Eq.~\ref{eq_large_scale_update}\;
 }
  Initialize $\Phi_{S_1}$ by ones, initialize $\Phi_{S_2}$ by zeros; Reset $\theta$ for few-shot tasks; Randomly initialize $\theta$; Initialize $\{m^*\}$ as an empty set.
 
 \For{iterations in meta-training}{
 Randomly sample a batch of tasks $\{\mathcal{T}_{1\sim K}\} \in p(\mathcal{T})$\;
%  Sample task $\mathcal{T}_i \in \{\mathcal{T}$\};
\For{$k$ from $1$ to $K$}{
 Optimize $\Phi_{S_{\{1,2\}}}$ and $\theta$ with $\mathcal{T}_k$ by \textbf{Algorithm}~\ref{alg_Meta}\;
 Get the returned the $m^*$-th class, then add it to a hard class set $\{m*\}$\;
 }
 Sample hard tasks $\{\mathcal{T}^{hard}\} \subseteq p(\mathcal{T}|\{m^*\})$\;
  \For{$k$ from $1$ to $K'$}{
 Sample task $\mathcal{T}^{hard}_k \in \{\mathcal{T}^{hard}$\} \;
 Optimize $\Phi_{S_{\{1,2\}}}$ and $\theta$ with $\mathcal{T}^{hard}_k$ by \textbf{Algorithm}~\ref{alg_Meta} \;
}
 Empty $\{m^*\}$.
}
\end{algorithm}
%%%%%%%%%%%%%%%%%%%%%%%%%%%%%%%%%%%%%%%%%%%%%%%%%%%%%%%%%%%

Algorithm~\ref{alg_Meta} presents an entire learning epoch 
on a single task. 
Base-learning steps in the episode training split are given in line 2-5. 
The meta-level update by the test loss is shown in line 6.
In line 7-11, the recognition rates of all test classes are computed and returned to Algorithm~\ref{alg_overall} (line 11) for hard task sampling.

Algorithm~\ref{alg_Meta_test} presents the evaluation process on unseen tasks. The average accuracy $Acc$ obtained in this Algorithm is used as the final evaluation of our method.

%%%%%%%%%%%%%%%%%%%%%%%%%%%%%%%%%%%%%%%%%%%%%%%%%%%%%%%%%%%

\begin{algorithm}
\caption{Detail learning steps of a training task}
\label{alg_Meta}
\SetAlgoLined
\SetKwInput{KwData}{Input}
\SetKwInput{KwResult}{Output}
 \KwData{Task $\mathcal{T}$, learning rates $ \beta$ and $\gamma$, feature extractor $\Theta$, base learner $\theta$, \emph{Scaling} and \emph{Shifting} parameters $\Phi_{S_{\{1,2\}}}$}
 \KwResult{Base learner $\theta$, \emph{Scaling} and \emph{Shifting} parameters $\Phi_{S_{\{1,2\}}}$, 
 the worst classified class-$m$ in $\mathcal{T}$}
 Sample training datapoints $\mathcal{T}^{(tr)}$ and test datapoints $\mathcal{T}^{(te)}$ from $\mathcal{T}$ \;
 \For{samples in $\mathcal{T}^{(tr)}$}{
 Evaluate $\mathcal{L}_{\mathcal{T}^{(tr)}}$\;
 Optimize $\theta'$ by Eq.~\ref{eq_base_classifier}\;
 }
 Optimize $\Phi_{S_{\{1,2\}}}$ and $\theta$ by Eq.~\ref{eq_ss_update} and Eq.~\ref{eq_meta_classifier}\;
 \For{$m \in \{1 \sim M\}$}{
 Classify samples of $m$-th class in $\mathcal{T}^{(te)}$\;
 Compute $Acc_m$\;
 }
 Return the $m^*$-th class with the lowest accuracy $Acc_m^*$.
\end{algorithm}

% $\theta'_K$
% $\theta'_1$
%%%%%%%%%%%%%%%%%%%%%%%%%%%%%%%%%%%%%%%%%%%%%%%%%%%%%%%%%%%

\begin{algorithm}
\caption{Meta test on the unseen task}
\label{alg_Meta_test}
\SetAlgoLined
\SetKwInput{KwData}{Input}
\SetKwInput{KwResult}{Output}
 \KwData{Unseen task $\mathcal{T}_{un}$, base learning rate $\beta$ and meta learning rate $\gamma$, 
 feature extractor $\Theta$, base learner $\theta$, \emph{Scaling} and \emph{Shifting} parameters $\Phi_{S_{\{1,2\}}}$}
 \KwResult{Average accuracy $Acc$}
 Sample training datapoints $\mathcal{T}^{(tr)}_{un}$ and test datapoints $\mathcal{T}^{(te)}_{un}$ from $\mathcal{T}_{un}$ \;
 \For{samples in $\mathcal{T}^{(tr)}_{un}$}{
 Evaluate $\mathcal{L}_{\mathcal{T}^{(tr)}_{un}}$\;
 Optimize $\theta'$ by Eq.~\ref{eq_base_classifier}\;
 }
  \For{samples in $\mathcal{T}^{(te)}_{un}$}{
 Predict class labels by $[\Theta; \theta']$\;
 Obtain the classification result\;
 }
 Compute the accuracy $Acc_m$ of $m$-th class\;
 Compute the mean accuracy $Acc$ of $M$ classes.
\end{algorithm}
%%%%%%%%%%%%%%%%%%%%%%%%%%

\section{Experiments}
\label{sec_exp}

We evaluate the proposed \textbf{MTL} and \textbf{HT meta-batch} in terms of few-shot recognition accuracy and model convergence speed.
Comparing to our conference version~\cite{SunCVPR2019}, this paper contains additional experiments on
(i) a larger and more challenging dataset -- tieredImageNet~\cite{RenICLR2018_semisupervised} which have been widely adopted in recent related works~\cite{RusuICLR2019, LeeCVPR19svm, LiuICLR2019transductive, LiCVPR2019bestResult};
(ii) two deeper architectures -- Resnet-18 and ResNet-25 as the Feature Extractor $\Theta$; and
(iii) supervised and semi-supervised MTL variances in which we implement \emph{SS} operations on both classical and state-of-the-art meta-learners, i.e. ProtoNets~\cite{SnellSZ17}, MatchingNets~\cite{VinyalsBLKW16}, RelationNets~\cite{SungCVPR2018}, and the meta-learning version of Baseline++ (superior to original Baseline++~\cite{chen19closerfewshot}), as well as the state-of-the-art semi-supervised few-shot classification models, i.e. Soft Masked $k$-Means~\cite{RenICLR2018_semisupervised} and TPN~\cite{LiuICLR2019transductive}.

Below we describe the datasets we evaluate on and detailed settings, followed by the comparisons to state-of-the-art methods, extensive validations on multiple MTL variances, and an ablation study regarding \emph{SS} operations and HT meta-batch.

\subsection{Datasets}
\label{sec_dataset}

We conduct few-shot learning experiments on three benchmarks, miniImageNet~\cite{VinyalsBLKW16}, tieredImageNet~\cite{RenICLR2018_semisupervised} and Fewshot-CIFAR100 (FC100)~\cite{OreshkinNIPS18}. 
miniImageNet is the most widely used in related works~\cite{FinnAL17, RaviICLR2017, GrantICLR2018, FranceschiICML18, MunkhdalaiICML18}, and the later ones are more recently published with a larger scale and a more challenging setting, i.e., lower image resolution and stricter training-test splits.

\myparagraph{miniImageNet~\cite{VinyalsBLKW16}.}
It was proposed especially for few-shot learning evaluation. 
Its complexity is high due to the use of ImageNet images, but it requires less resource and infrastructure than running on the full ImageNet dataset~\cite{Russakovsky2015}. 
In total, there are $100$ classes with $600$ samples of $84 \times 84$ color images per class.
These $100$ classes are divided into $64$, $16$, and $20$ classes respectively for sampling tasks for meta-training, meta-validation and meta-test, following related works~\cite{FinnAL17, RaviICLR2017, GrantICLR2018, FranceschiICML18, MunkhdalaiICML18}.

\myparagraph{tieredImageNet~\cite{RenICLR2018_semisupervised}.}
Compared to miniImageNet, it is a larger subset of ImageNet with $608$ classes ($779,165$ images) grouped into $34$ super-class nodes. These nodes are partitioned into $20$, $6$, and $8$ disjoint sets respectively for meta- training, validation, and test. The corresponding sub-classes are used to build the classification tasks in each of which the $5$ sub-classes are randomly sampled. 
As argued in~\cite{RenICLR2018_semisupervised}, this super-class based training-test split results in a more challenging and realistic regime with meta- test and validation tasks that are less similar to meta-training tasks.

\myparagraph{Fewshot-CIFAR100 (FC100)~\cite{OreshkinNIPS18}.}
This dataset is based on the popular object classification dataset CIFAR100~\cite{CIFAR100}. Its training-test splits are also based on super-classes~\cite{OreshkinNIPS18}.
In total, it contains $100$ object classes ($600$ images per class) belonging to $20$ super-classes. Meta-training data are from $60$ classes belonging to $12$ super-classes. Meta-validation and meta-test sets contain $20$ classes belonging to $4$ super-classes, respectively. 
Comparing to the ImageNet subsets above, FC100 offers a more challenging scenario with lower image resolution, i.e. each sample is a $32 \times 32$ color image. 
%

%%%%%%%%%%%%%%%%%%%%%% shared settings as follows,
\subsection{Implementation details}

\myparagraph{Task sampling.} 
We use the same task sampling method as related works~\cite{FinnAL17}, on all datasets. Specifically, (1) we consider the 5-class classification, (2) during meta-training, we sample 5-class, 1-shot (or 5-shot) episodes to contain $1$ (or $5$) samples for train episode and $15$ (uniform) samples for episode test, and (3) during meta-validation and meta-test, we sample 5-class, 1-shot (or 5-shot) episodes to contain $1$ (or $5$) samples for train episode and $1$ (uniform) sample for episode test.
Note that in some related works, e.g. ~\cite{OreshkinNIPS18}, $32$ samples are used for episode test on 5-shot tasks. Using such a larger number of test samples results in the lower standard variance of recognition accuracies.

In total, we sample at most $20k$ tasks ($10k$ meta-batches) for meta-training (same for the cases \emph{w/} and \emph{w/o} HT meta-batch), and sample $600$ random tasks for both meta-validation and meta-test~\cite{FinnAL17}. 
Note that we choose the trained models which have the highest meta-validation accuracies, for meta-test. 

\myparagraph{Network architectures.} 
We present the details of network architectures for
Feature Extractor parameters $\Theta$, 
MTL meta-learner with \emph{Scaling} and \emph{Shifting} parameters $\Phi_{S_1}, \Phi_{S_2}$, and MTL base-learner (classifier) parameters $\theta$. 
For $\Theta$,
in our conference version~\cite{SunCVPR2019}, we used ResNet-12 and 4CONV which have been commonly used in previous works~\cite{FinnAL17, VinyalsBLKW16, RaviICLR2017, MunkhdalaiICML18, MishraICLR2018, OreshkinNIPS18, LeeCVPR19svm}.
In this journal version, we implement two deeper architectures -- ResNet-18 and ResNet-25, which have been adopted in newly published related works~\cite{LiCVPR2019bestResult, QiaoCVPR2018, YeArXiv2018}, and we achieve the top performance using ResNet-25.

Specifically, \textbf{4CONV} consists of $4$ layers with $3\times 3$ convolutions and $32$ filters, followed by batch normalization (BN)~\cite{IoffeICML15}, a ReLU nonlinearity, and $2\times 2$ max-pooling.
MTL only works with the following deep nets. \textbf{ResNet-12} contains $4$ residual blocks and each block has $3$ CONV layers with $3\times 3$ kernels.
At the end of each residual block, a $2\times 2$ max-pooling layer is applied. The number of filters starts from $64$ and is doubled every next block. 
Following $4$ blocks, there is a mean-pooling layer to compress the output feature maps to a feature embedding.
\textbf{ResNet-18} contains $4$ residual blocks and each block has $4$ CONV layers with $3\times 3$ kernels. The number of filters starts from $64$ and is doubled every next block. Before the residual blocks, there is one additional CONV layer with $64$ filter and $3\times 3$ kernels at the beginning of the network. The residual blocks are followed by an average pooling layer. The ResNet-18 backbone we use exactly follow~\cite{HeZRS16} except that the last FC layer is removed.
\textbf{ResNet-25} is exactly the same as the released code of~\cite{QiaoCVPR2018, ye2018learning}. Three residual blocks are used after an initial convolutional layer. Each block has $4$ CONV layers with $3\times 3$ kernels. The number of filters starts from $160$ and is doubled every next block. After a global average pooling layer, it leads to a $640$-dim embedding.
For fair comparison with important baseline methods such as MAML~\cite{FinnAL17} and ProtoNets~\cite{SnellSZ17} which are based on 4CONV nets, we implement the experiments for them using the same ResNet-12 and ResNet-18 architectures.

For the architecture of $\Phi_{S_1}$ and $\Phi_{S_2}$, actually, they are generated according to the architecture of $\Theta$, as introduced in Section~\ref{sec_meta_transfer}. For example, when using ResNet-25 in MTL, $\Phi_{S_1}$ and $\Phi_{S_2}$ also have 25 layers, respectively.

For the architecture of $\theta$ (the parameters of the base-learner),
we empirically find that in our cases a single FC layer (as $\theta$) is faster to train and more effective for classification than multiple layers, taking the most popular dataset miniImageNet as an example. 
Results are given in Table~\ref{table_diff_fc_arch}, in which we can see the performance drop when changing this $\theta$ to multiple layers.

\myparagraph{Pre-training stage.}
For the phase of DNN training on large-scale data (Section~\ref{sec_large_scale_pretrain}), the model is trained by Adam optimizer~\cite{kingma2014adam}. The learning rate is initialized as $0.001$, and decays to its half every $5k$ iterations until it is lower than $0.0001$. 
We set the keep probability of the dropout as $0.9$ and batch-size as $64$.
The pre-training stops after $10k$ iterations.
Note that for hyperparameter selection, we randomly choose $550$ samples each class as the training set, and the rest as validation. 
After the grid search for hyperparameters, we fix them and mix up all samples ($64$ classes, $600$ samples each class) to do the final pre-training.
Image samples in these steps are augmented by horizontal flipping.

\myparagraph{Meta-training stage.}
This is a task-level training in which the base-learning in one task considers a training step for optimizing base-learner, followed by a validation step for optimizing meta-learner.
The base-learner $\theta$ is optimized by batch gradient descent with the learning rate of $0.01$. It is updated with $20$ and $60$ epochs respectively for 1-shot and 5-shot tasks on the miniImageNet and tieredImageNet datasets, and $20$ epochs for all tasks on the FC100 dataset. Specially when using ResNet-25, we use 100 epochs for all tasks on all datasets.
The meta-learner, i.e. the parameters of the \emph{SS} operations, is optimized by Adam optimizer~\cite{kingma2014adam}.
Its learning rate is initialized as $0.001$, and decays to the half every $1k$ iterations until $0.0001$. 
The size of meta-batch is set to $2$ (tasks) due to the memory limit.

\myparagraph{HT meta-batch.}
Hard tasks are sampled every time after running $10$ meta-batches, i.e., the failure classes used for sampling hard tasks are from $20$ tasks as each meta-batch contains $2$ tasks. The number of hard tasks is selected for different settings by validation: $10$ and $4$ hard tasks respectively for the 1-shot and 5-shot experiments, on the miniImageNet and tieredImageNet datasets; and respectively $20$ and $10$ hard tasks for the 1-shot, 5-shot experiments, on the FC100 dataset.

\begin{table}[htp]%\scriptsize
  \small
  \centering
  \begin{tabular}{clcccc}
    \toprule
 \multirow{2}{*}{Base-learning} & \multirow{2}{*}{Dim. of $\theta$}  & \multicolumn{2}{c}{miniImageNet}\\
       & & 1-shot & 5-shot \\
% $\theta$ & 5 & 4 CONV (pre) & 45.6\tiny{$\pm$ $1.8$} & 61.2 \tiny{$\pm$ $0.9$} \\
     \midrule[0.7pt]
$\theta$ (2 FC layers) & 512, 5& 59.1 \tiny{$\pm$ $1.9$} & 70.7 \tiny{$\pm$ $0.9$} \\
$\theta$ (3 FC layers) & 1024, 512, 5 & 56.2 \tiny{$\pm$ $1.8$} & 68.7 \tiny{$\pm$ $0.9$}\\
     \midrule
$\Theta$, $\theta$ & 5 & 59.6 \tiny{$\pm$ $1.8$} & 71.6 \tiny{$\pm$ $0.9$} \\     
    \midrule
$\theta$ (\textbf{Ours}) & 5 & \textbf{60.6 \tiny{$\pm$ $1.9$}} & \textbf{74.3 \tiny{$\pm$ $0.8$}} \\
  \bottomrule
\end{tabular}
  \vspace{0.1cm}
  \caption{The 5-way, 1-shot and 5-shot classification accuracy ($\%$) on miniImageNet, when using different architectures for the base-learner (i.e., the classifier $\theta$).}
    \label{table_diff_fc_arch}
\end{table}

\subsection{Ablation setting}
\label{sec_setting_ablation}

In order to show the effectiveness of our approach, we carefully design several ablative settings: 
two baselines without meta-learning but more classic learning, named as \emph{update*},
four baselines of \emph{Fine-Tuning} (\emph{FT}) on different numbers of parameters in the outer-loop based on MAML~\cite{FinnAL17}, named as \emph{FT*}, and
two \emph{SS} variants on smaller numbers of parameters, named as \emph{SS*}.
Table~\ref{table_ablation} shows the results in these settings, for which we simply use the classical architecture (ResNet-12) containing $4$ residual blocks named $\Theta 1 \sim \Theta4$ and an FC layer $\theta$ (classifier).
The bullet names used in the Table are explained as follows.

\myparagraph{\emph{update} $[\Theta; \theta]$ (or $\theta$).} 
There is no meta-training phase. During test phase, each task has its whole model $[\Theta; \theta]$ (or the classifier $\theta$) updated on $\mathcal{T}^{(tr)}$, and then tested on $\mathcal{T}^{(te)}$.

\myparagraph{\emph{FT} $\theta$ ($[\Theta 4; \theta]$ or $[\Theta 3; \Theta 4; \theta]$ or $\theta$).}
These are straight-forward ways to reduce the quantity of meta-learned parameters. For example, ``$[\Theta 3; \Theta 4; \theta]$'' does not update the the first two residual blocks which encode the low-level image features. Specially, ``$\theta$'' means only the classifier parameters are updated during meta-training.

\myparagraph{\emph{SS} $[\Theta 4; \theta]$ (or $[\Theta 3; \Theta 4; \theta]$ or $\theta$).}
During the meta-training, a limited number of \emph{SS} parameters are defined and used. Low-level residual blocks simply use the pre-trained weights without meta-level update.

\subsection{Comparison with related works}
\label{sec_overall_results}

\begin{table*}
  \small
  \centering
  \begin{tabular}{l l lcc}
    \toprule
     \multicolumn{2}{c}{\multirow{2}{*}{\textbf{Few-shot Learning Method}}} & \multirow{2}{*}{\textbf{Backbone}} &  \multicolumn{2}{c}{\textbf{miniImageNet (test)}} \\
     \cmidrule{4-5}
     &&& 1-shot & 5-shot \\
    \midrule
    \multirow{2}{*}{\emph{Data augmentation}}
    & Adv. ResNet, \cite{Mehrotra2017} & WRN-40 (pre) & 55.2 & 69.6 \\
    & Delta-encoder, \cite{SchwartzNIPS18} & VGG-16 (pre) & 58.7 & 73.6 \\
    \midrule  
    \multirow{6}{*}{\emph{Metric learning}}
    &MatchingNets, \cite{VinyalsBLKW16} & 4 CONV & 43.44 $\pm$ $0.77$ & 55.31 $\pm$  $0.73$ \\
    &ProtoNets, \cite{SnellSZ17} & 4 CONV & 49.42 $\pm$ $0.78$ & 68.20 $\pm$ $0.66$\\
    &RelationNets, \cite{SungCVPR2018} & 4 CONV & 50.44 $\pm$ $0.82$ & 65.32 $\pm$ $0.70$\\
    &Graph neural network, \cite{SatorrasICLR2018graph} & 4 CONV & 50.33 $\pm$ $0.36$ & 66.41 $\pm$ $0.63$ \\
    &Ridge regression, \cite{BertinettoICLR2019ridge} & 4 CONV & 51.9 $\pm$ $0.2$ & 68.7$\pm$ $0.2$ \\
    &TransductiveProp, \cite{LiuICLR2019transductive} & 4 CONV & 55.51 & 69.86 \\
    \midrule
    \multirow{3}{*}{\emph{Memory network}} 
    % & Neural Turing Machines~\cite{SantoroBBWL16} no results on mini
    & Meta Networks, \cite{MunkhdalaiICML2017} & 5 CONV  & 49.21 $\pm$ $0.96$ & -- \\
    & SNAIL, \cite{MishraICLR2018} & ResNet-12 (pre)${}^{\diamond}$  & 55.71 $\pm$ $0.99$  & 68.88 $\pm$ $0.92$\\
    & TADAM, \cite{OreshkinNIPS18} & ResNet-12 (pre)${}^{\dag}$  & 58.5 $\pm$ $0.3$  & 76.7 $\mathbf{\pm}$ $\mathbf{0.3}$\\
    \midrule
    \multirow{10}{*}{\emph{Gradient descent}}
    & MAML, \cite{FinnAL17} & 4 CONV & 48.70 $\pm$ $1.75$ & 63.11 $\pm$ $0.92$ \\
    & Meta-LSTM, \cite{RaviICLR2017} & 4 CONV & 43.56 $\pm$ $0.84$ & 60.60 $\pm$ $0.71$ \\
    & Hierarchical Bayes, \cite{GrantICLR2018} & 4 CONV  & 49.40 $\pm$ $1.83$ & -- \\
    & Bilevel Programming, \cite{FranceschiICML18} & ResNet-12${}^{\diamond}$   & 50.54 $\pm$ $0.85$  & 64.53 $\pm$ $0.68$\\
    & MetaGAN, \cite{ZhangNIPS2018MetaGAN} & ResNet-12 & 52.71 $\pm$ $0.64$  & 68.63 $\pm$ $0.67$ \\
    & adaResNet, \cite{MunkhdalaiICML18} & ResNet-12${}^{\ddag}$   & 56.88 $\pm$ $0.62$ & 71.94 $\pm$ $0.57$ \\
    & MetaOptNet, \cite{LeeCVPR19svm} & ResNet-12 & 62.64 $\pm$ $0.35$ & 78.63 $\pm$ $0.68$ \\
    & LEO, \cite{RusuICLR2019} & WRN-28-10 (pre) & 61.67 $\pm$ $0.08$ & 77.59 $\pm$ $0.12$ \\
    & LGM-Net, \cite{LiICML2019best1shotResult} & MetaNet+4CONV & \textbf{69.13} $\pm$ $0.35$ & 71.18 $\pm$ $0.68$ \\
    & CTM, \cite{LiCVPR2019bestResult} & ResNet-18 (pre) & 64.12 $\pm$ $0.82$ & 80.51 $\pm$ $0.13$ \\
    % 64.12 ± 0.82 80.51 ± 0.13
    % 69.13±0.35% 71.18±0.68%
    \midrule
    % MAML, HT  & \emph{FT} $[\Theta; \theta]$, HT meta-batch & 4 CONV & 49.1 $\pm$ $1.9$ & 64.1 $\pm$ $0.9$ \\
    \multirow{4}{*}{\textbf{Ours}}
    & \emph{FT} $[\Theta; \theta]$, HT meta-batch & ResNet-12 (pre) & 59.1 $\pm$ $1.9$ & 73.1 $\pm$ $0.9$ \\
    % & \emph{SS} $[\Theta; \theta]$, meta-batch & ResNet-12 (pre) & 60.2  $\pm$ $1.9$& 74.3 $\pm$ $0.9$ \\
    & \emph{SS} $[\Theta; \theta]$, HT meta-batch & ResNet-12 (pre) & 61.2 $\pm$ $1.8$ & 75.5 $\pm$ $0.8$\\
    % & \emph{SS} $[\Theta; \theta]$, meta-batch & ResNet-18 (pre) & 60.8 $\pm$ $1.9$ & 74.5$\pm$ $0.8$\\
    & \emph{SS} $[\Theta; \theta]$, HT meta-batch & ResNet-18 (pre) & 61.7 $\pm$ $1.8 $& 75.6 $\pm$ $0.9$\\
        % & \emph{SS} $[\Theta; \theta]$, meta-batch & ResNet-25 (pre) & 63.4 $\pm$ $1.8$ & 80.1 $\pm$ $0.9$\\
    & \emph{SS} $[\Theta; \theta]$, HT meta-batch & ResNet-25 (pre) & 64.3 $\pm$ $1.7$ & \textbf{80.9} $\pm$ $0.8$\\
    \midrule[0.7pt]

    \multicolumn{2}{c}{\multirow{2}{*}{\textbf{Few-shot Learning Method}}} & \multirow{2}{*}{\textbf{Backbone}} &  \multicolumn{2}{c}{\textbf{tieredImageNet (test)}} \\
    \cmidrule{4-5}
     &&& 1-shot & 5-shot \\
    \midrule
    \multirow{3}{*}{\emph{Metric learning}}
    &ProtoNets,~\cite{SnellSZ17} (by \cite{RenICLR2018_semisupervised}) & 4 CONV & 53.31 $\pm$ $0.89$ & 72.69 $\pm$ $0.74$\\
    &RelationNets, \cite{SungCVPR2018} (by \cite{LiuICLR2019transductive}) & 4 CONV & 54.48 $\pm$ $0.93$ & 71.32 $\pm$ $0.78$\\
    &TransductiveProp, \cite{LiuICLR2019transductive} & 4 CONV & 57.41 $\pm$ $0.94$ & 71.55 $\pm$ $0.74$ \\
    \midrule
    \multirow{3}{*}{\emph{Gradient descent}}
    & MAML, \cite{FinnAL17} (by~\cite{LiuICLR2019transductive}) & ResNet-12 & 51.67 $\pm$ $1.81$ & 70.30 $\pm$ $0.08$ \\
    & LEO, \cite{RusuICLR2019} & WRN-28-10 (pre) & 66.33 $\pm$ $0.05$ & 81.44 $\pm$ $0.09$ \\
    & CTM, \cite{LiCVPR2019bestResult} & ResNet-18 (pre) & 68.41 $\pm$ $0.39$ & 84.28 $\pm$ $1.73$ \\
    % 68.41 ± 0.39 84.28 ± 1.73
    \midrule
    %   MAML, HT  & \emph{FT} $[\Theta; \theta]$, HT meta-batch & 4 CONV & w & w  \\
    
    \multirow{4}{*}{\textbf{Ours}}
    &  \emph{FT} $[\Theta; \theta]$, HT meta-batch & ResNet-12 (pre) & 64.8 $\pm$ $1.9$ & 78.4 $\pm$ $0.9$ \\
    % & \emph{SS} $[\Theta; \theta]$, meta-batch & ResNet-12 (pre) & 65.6 $\pm$ $1.8$ & 80.6 $\pm$ $0.9$ \\
    & \emph{SS} $[\Theta; \theta]$, HT meta-batch & ResNet-12 (pre) & 65.6 $\pm$ $1.8$ & 80.8 $\pm$ $0.8$\\
    %  & \emph{SS} $[\Theta; \theta]$, meta-batch & ResNet-18 (pre) & 67.1 $\pm$ $1.8$ & 82.1 $\pm$ $0.9$\\
     & \emph{SS} $[\Theta; \theta]$, HT meta-batch & ResNet-18 (pre) & 67.5 $\pm$ $1.9$ & 82.5 $\pm$ $0.7$\\
        %  & \emph{SS} $[\Theta; \theta]$, meta-batch & ResNet-25 (pre) & 69.1 $\pm$ $1.7$ & 84.2 $\pm$ $0.8$\\
    & \emph{SS} $[\Theta; \theta]$, HT meta-batch & ResNet-25 (pre) & \textbf{72.0} $\pm$ $1.8$ & \textbf{85.1} $\pm$ $0.8$\\
     \midrule[0.7pt]

 \multicolumn{2}{c}{\multirow{2}{*}{\textbf{Few-shot Learning Method}}} & \multirow{2}{*}{\textbf{Backbone}} &  \multicolumn{2}{c}{\textbf{FC100 (test)}} \\
 \cmidrule{4-5}
     &&& 1-shot & 5-shot\\
    \midrule    
    \multirow{3}{*}{\emph{Gradient descent}} 
    & MAML, ~\cite{FinnAL17} (by us) & 4 CONV  &  38.1 $\pm$ $1.7$ & 50.4 $\pm$ $1.0$\\
    & MAML++, ~\cite{AntoniouICLR19} (by us) & 4 CONV  &  38.7 $\pm$ $0.4$ & 52.9 $\pm$ $0.4$\\
    & MetaOptNet~\cite{LeeCVPR19svm} & 4 CONV  &  41.1 $\pm$ $0.6$ & 55.5 $\pm$ $0.6$\\
    \midrule
    \multirow{1}{*}{\emph{Memory network}}
    & TADAM, \cite{OreshkinNIPS18} & ResNet-12 (pre)${}^{\dag}$   & 40.1 $\pm$ $0.4$  & 56.1 $\pm$ $0.4$ \\
    \midrule
%   MAML, HT  & \emph{FT} $[\Theta; \theta]$, HT meta-batch & 4 CONV & 39.9 $\pm$ $1.8$ & 51.7 $\pm$ $0.9$  \\
    \multirow{2}{*}{\emph{Metric learning}}
    &MatchingNets,~\cite{VinyalsBLKW16} (by us) & ResNet-25 (pre) & 44.2 $\pm$ $1.8$ & 58.0 $\pm$ $0.8$\\
    &RelationNets, \cite{SungCVPR2018} (by us) & ResNet-25 (pre) & 41.1 $\pm$ $1.8$ & 58.6 $\pm$ $0.8$\\
    \midrule
    \multirow{4}{*}{\textbf{Ours}}
    &  \emph{FT} $[\Theta; \theta]$, HT meta-batch & ResNet-12 (pre) & 41.8 $\pm$ $1.9$ & 55.1 $\pm$ $0.9$ \\
    % & \emph{SS} $[\Theta; \theta]$, meta-batch & ResNet-12 (pre) & 43.6 $\pm$ $1.8$ & 55.4 $\pm$ $0.9$\\
    & \emph{SS} $[\Theta; \theta]$, HT meta-batch & ResNet-12 (pre) & 45.1 $\pm$ $1.9$ & 57.6 $\pm$ $0.9$\\
    % & \emph{SS} $[\Theta; \theta]$, meta-batch & ResNet-18 (pre) & 44.7 $\pm$ $1.9$ & 57.5 $\pm$ $1.0$\\
    & \emph{SS} $[\Theta; \theta]$, HT meta-batch & ResNet-18 (pre) & 45.5 $\pm$ $1.9$ &  57.6 $\pm$ $1.0$\\
    % & \emph{SS} $[\Theta; \theta]$, meta-batch & ResNet-25 (pre) & 45.6 $\pm$ $1.8$ & 60.8 $\pm$ $0.9$\\
    & \emph{SS} $[\Theta; \theta]$, HT meta-batch & ResNet-25 (pre) & \textbf{46.0} $\pm$ $1.7$ & \textbf{61.3} $\pm$ $0.8$\\
  \bottomrule
    \multicolumn{5}{l}{${}^{\diamond}$Additional 2 convolutional layers { } ${}^{\ddag}$Additional 1 convolutional layer { } ${}^{\dag}$Additional 72 fully connected layers}\\
\end{tabular}
  \vspace{0.1cm}
  \caption{The 5-way, 1-shot and 5-shot classification accuracy ($\%$) on miniImageNet, tieredImageNet, and FC100 datasets.
``pre'' means including our pre-training step with all training datapoints. ``by [*]'' means the results were reported in [*]. ``by us'' means the results are from our implementation of open-sourced code. Note that the standard variance is affected by the number of episode test samples, and our sample splits are the same with MAML~\cite{FinnAL17}.}
    \label{table_sota_all}
\end{table*}

Table~\ref{table_sota_all} presents the overall comparisons with related works, on the miniImageNet, tieredImageNet, and FC100 datasets. 
Note that these numbers are the meta-test results of the meta-trained models which have the highest meta-validation accuracies.
Specifically, on the miniImageNet, models on 1-shot and 5-shot are meta-trained for $6k$ and $10k$ iterations, respectively. 
On the tieredImageNet, iterations for 1-shot and 5-shot are at $8k$ and $10k$, respectively.
On the FC100, iterations are all at $3k$.

\myparagraph{miniImageNet.} 
In the first block of Table~\ref{table_sota_all}, we can see that the ResNet-25 based MTL model learned in HT meta-batch scheme achieves the best few-shot classification performance, i.e. $80.9\%$, for (5-class, 5-shot). 
It tackles the $1$-shot tasks with an accuracy of $64.3\%$, comparable to the state-of-the-art result ($69.13\%$) reported by LGM-Net~\cite{LiICML2019best1shotResult} whose $5$-shot result is $9.7\%$ lower than ours.
Regarding the network architecture, we can see that
models using deeper ones, i.e. ResNet-12, ResNet-18, and ResNet-25, outperforms the 4CONV based models by rather large margins, e.g. 4CONV models have the best 1-shot result with $55.51\%$~\cite{LiuICLR2019transductive} which is $8.8\%$ lower than our $64.3\%$. This demonstrates our contribution of utilizing deeper neural networks to better tackle the few-shot classification problems.

\myparagraph{tieredImageNet.}
In the second block of Table~\ref{table_sota_all},  we give the results on the large-scale dataset -- tieredImageNet. Since this dataset is newly proposed~\cite{RenICLR2018_semisupervised}, its results using classical models~\cite{SnellSZ17, SungCVPR2018, FinnAL17} were reported in recent papers~\cite{RenICLR2018_semisupervised, LiuICLR2019transductive}.
From the table, we can see that the ResNet-25 MTL outperforms others, e.g. it achieves around a $4\%$ margin over the top-performed CTM~\cite{LiCVPR2019bestResult} on 1-shot tasks.
An interesting observation is that on this larger and more challenging dataset, our deeper version of MTL (with ResNet-25) outperforms the shallower one (with ResNet-12) by $6.4\%$ on 1-shot, which is the double of the improvement ($3.2\%$) obtained on miniImageNet. This demonstrates that our idea of using deeper neural networks shows to be more promising for handling more difficult few-shot learning scenarios.

\myparagraph{FC100.}
In the last block of Table~\ref{table_sota_all}, we show the results on the FC100. We report the results of TADAM~\cite{OreshkinNIPS18} and MetaOptNet~\cite{LeeCVPR19svm} using the numbers given in original papers, and obtain the results of classical methods -- MAML~\cite{FinnAL17}, MAML++~\cite{AntoniouICLR19}, RelationNets~\cite{SungCVPR2018} and MatchineNets~\cite{VinyalsBLKW16} by implementing their open-sourced code with our deep networks.
From these results, we can see that MTL consistently outperforms the original MAML and the improved version -- MAML++ by large margins, e.g. around $7\%$ in the 1-shot cases. It also surpasses both TADAM and MetaOptNet by about $5\%$. 

%%%%%%%%%%%%%%%%%%%%%%%%%%%%%%%%%%%%%%%%%%%%%%%%%%%%%%%

From the overview of Table~\ref{table_sota_all}, we can conclude that our large-scale network pre-training and transferring on deeper CNNs significantly boost the few-shot learning performance. 
In this journal version, we add experiments on the tieredImageNet dataset whose meta-train and meta-test are clearly split according to super-classes.
It is interesting to observe that our performance improvements on this more challenging dataset is clearly more than those on miniImageNet, and this is more obvious on the 1-shot tasks for which our best model achieves $3.6\%$ higher than the new state-of-the-art model -- CTM~\cite{LiCVPR2019bestResult}.

\subsection{Generalization ability of MTL}
\label{sec_generalization}

\begin{table*}
%   \vspace{0.2cm}
  \small
  \centering
  \begin{tabular}{clccccccccc}
    \toprule
     \multirow{2}{*}{Backbone} & \multirow{2}{*}{Method} & \multirow{2}{*}{Operation} &  \multicolumn{2}{c}{miniImageNet} & & \multicolumn{2}{c}{miniImageNet (tieredPre)} & & \multicolumn{2}{c}{tieredImageNet} \\
     \cmidrule{4-5}\cmidrule{7-8}\cmidrule{10-11}
     & & & 1-shot & 5-shot && 1-shot & 5-shot && 1-shot & 5-shot \\
    \midrule[0.7pt]
    \multirow{12}{*}{ResNet-12}

    & \multirow{2}{*}{ProtoNets~\cite{SnellSZ17}} & \emph{SS} & 56.7 \tiny{$\pm$ 1.9} & 72.0 \tiny{$\pm$ 0.9} && 62.0 \tiny{$\pm$ 1.9} & 77.9 \tiny{$\pm$ 1.0} && 62.2 \tiny{$\pm$ 2.1} & 78.1 \tiny{$\pm$ 0.9} \\
    & & \emph{FT}   & 55.2 \tiny{$\pm$ 1.9} & 70.8 \tiny{$\pm$ 0.9} && 57.2 \tiny{$\pm$ 1.9} & 75.9 \tiny{$\pm$ 0.9} && 54.9 \tiny{$\pm$ 2.0} & 73.0 \tiny{$\pm$ 1.0}\\
    \cmidrule{2-11}
    & \multirow{2}{*}{MatchingNets~\cite{VinyalsBLKW16}} & \emph{SS}   & 58.1 \tiny{$\pm$ 1.8} & 66.9 \tiny{$\pm$ 0.9} && 63.6 \tiny{$\pm$ 1.7} & 73.2 \tiny{$\pm$ 0.9} && 64.5 \tiny{$\pm$ 1.9} & 73.9 \tiny{$\pm$ 0.9}\\
    & & \emph{FT}   & 57.4 \tiny{$\pm$ 1.7} & 67.5 \tiny{$\pm$ 0.8} && 61.1 \tiny{$\pm$ 1.8} & 72.6 \tiny{$\pm$ 0.8} && 62.4 \tiny{$\pm$ 1.8} & 73.5 \tiny{$\pm$ 0.8}\\
    \cmidrule{2-11}
    & \multirow{2}{*}{RelationNets~\cite{SungCVPR2018}} & \emph{SS}   & 57.2 \tiny{$\pm$ 1.8} & 71.1 \tiny{$\pm$ 0.9} && 61.5 \tiny{$\pm$ 1.8} & 74.9 \tiny{$\pm$ 0.9} && 65.6 \tiny{$\pm$ 1.9} & 77.5 \tiny{$\pm$ 0.9}\\
    & & \emph{FT}   & 56.0 \tiny{$\pm$ 1.8} & 69.0 \tiny{$\pm$ 0.8} && 58.9 \tiny{$\pm$ 1.8} & 72.0 \tiny{$\pm$ 0.8} && 62.2 \tiny{$\pm$ 1.8} & 76.0 \tiny{$\pm$ 0.9}\\
    \cmidrule{2-11}
    
    & MTL (FC) & \emph{SS}  & \textbf{60.6} \tiny{\tiny{$\pm$ 1.9}} & 74.3 \tiny{$\pm$ 0.8} && 65.7 \tiny{\tiny{$\pm$ 1.8}} & 78.4 \tiny{$\pm$ 0.8} & & 65.6 \tiny{$\pm$ 1.7} & 78.7 \tiny{$\pm$ 0.9} \\
    & MAML~\cite{FinnAL17} (FC) & \emph{FT}   & 58.3 \tiny{$\pm$ 1.9} & 71.6 \tiny{$\pm$ 0.9} && 61.6 \tiny{$\pm$ 1.9} & 73.5 \tiny{$\pm$ 0.8} && 62.0 \tiny{$\pm$ 1.8} & 70.6 \tiny{$\pm$ 0.9} \\
    \cmidrule{2-11}
    % & \multirow{2}{*}{MetaOptNet-SVM~\cite{LeeCVPR19svm}} & \emph{SS}   &  & &&&&&&\\
    % & & \emph{FT}   & & &&&&&&\\
    & MTL (Cosine) & \emph{SS} & 58.2 \tiny{$\pm$ 1.8} & \textbf{74.6} \tiny{$\pm$ 0.8} && \textbf{66.1} \tiny{$\pm$ 1.8} & \textbf{79.7} \tiny{$\pm$ 0.9} && \textbf{67.1} \tiny{$\pm$ 1.8} & \textbf{80.0} \tiny{$\pm$ 0.8} \\
    & MAML~\cite{FinnAL17} (Cosine) & \emph{FT}   & 59.8 \tiny{$\pm$ 1.8} & 72.8 \tiny{$\pm$ 0.9}  && 59.9 \tiny{$\pm$ 1.9} & 76.5 \tiny{$\pm$ 0.7} & & 65.1 \tiny{$\pm$ 1.9} & 78.2 \tiny{$\pm$ 0.8} \\
    
    %%%%%%%%%%%%%%%%%%%%%%%%%%%%%
    
    \midrule[0.7pt]
    \multirow{12}{*}{ResNet-18}
    
    & \multirow{2}{*}{ProtoNets~\cite{SnellSZ17}} & \emph{SS}   & 55.6 \tiny{$\pm$ 1.9} & 70.7 \tiny{$\pm$ 0.9} && 63.0 $\pm$ w & 78.3 \tiny{$\pm$ 1.0} && 58.6 \tiny{$\pm$ 2.0} & 77.8\tiny{$\pm$ 1.0}\\
    & & \emph{FT}   & 55.5 \tiny{$\pm$ 1.8} & 70.4 \tiny{$\pm$ 0.8} && 59.2 \tiny{$\pm$ 1.8} & 74.6 \tiny{$\pm$ 0.8} && 58.4 \tiny{$\pm$ 1.7} & 75.0 \tiny{$\pm$ 0.8} \\
    \cmidrule{2-11}
    & \multirow{2}{*}{MatchingNets~\cite{VinyalsBLKW16}} & \emph{SS}   & 57.2 \tiny{$\pm$ 1.9} & 65.9 \tiny{$\pm$ 0.9} && 62.3 \tiny{$\pm$ 1.8} & 74.0 \tiny{$\pm$ 0.9} && 64.3 \tiny{$\pm$ 1.9} & 75.5 \tiny{$\pm$ 0.9}\\
    & & \emph{FT}   & 56.9 \tiny{$\pm$ 1.7} & 68.4 \tiny{$\pm$ 0.8} && 62.2 \tiny{$\pm$ 1.8} & 72.0 \tiny{$\pm$ 0.8} && 62.9 \tiny{$\pm$ 1.8} & 75.9 \tiny{$\pm$ 0.8}\\
    \cmidrule{2-11}
    & \multirow{2}{*}{RelationNets~\cite{SungCVPR2018}} & \emph{SS}   &  57.6 \tiny{$\pm$ 0.8} & 71.1 \tiny{$\pm$ 0.8}  && 60.7 \tiny{$\pm$ 1.9} & 74.6 \tiny{$\pm$ 1.9} &&  61.9 \tiny{$\pm$ 1.8} & 77.8 \tiny{$\pm$ 0.9}\\
    & & \emph{FT}   & 51.5 \tiny{$\pm$ 1.8} & 64.4 \tiny{$\pm$ 0.8} && 46.8 \tiny{$\pm$ 1.8} & 66.4 \tiny{$\pm$ 0.8} && 58.4 \tiny{$\pm$ 1.8} & 71.8 \tiny{$\pm$ 0.8} \\
    \cmidrule{2-11}
    
    & MTL (FC) & \emph{SS}   & \textbf{60.8} \tiny{$\pm$ 1.9} & \textbf{74.5} \tiny{$\pm$ 0.8} && 66.5 \tiny{$\pm$ 1.8} &  80.2 \tiny{$\pm$ 0.8} && \textbf{67.0} \tiny{$\pm$ 1.8} & 80.6 \tiny{$\pm$ 0.8} \\
    &  MAML~\cite{FinnAL17} (FC) & \emph{FT}   & 56.8 \tiny{$\pm$ 1.9} & 65.4 \tiny{$\pm$ 0.7} && 50.9 \tiny{$\pm$ 1.8} & 68.8 \tiny{$\pm$ 0.8} && 50.5 \tiny{$\pm$ 1.8} & 71.8 \tiny{$\pm$ 0.8} \\
    \cmidrule{2-11}
    &MTL (Cosine) & \emph{SS}   & 59.4 \tiny{$\pm$ 1.9}& 74.3 \tiny{$\pm$ 0.8} && \textbf{66.7} \tiny{$\pm$ 1.8} & \textbf{81.8} \tiny{$\pm$ 0.8} && 66.6 \tiny{$\pm$ 1.8} & \textbf{82.1} \tiny{$\pm$ 0.8} \\
    & MAML~\cite{FinnAL17} (Cosine) & \emph{FT}   & 58.1 \tiny{$\pm$ 1.9} & 69.9 \tiny{$\pm$ 0.8} && 62.1 \tiny{$\pm$ 1.9} & 77.7 \tiny{$\pm$ 0.8} && 64.8 \tiny{$\pm$ 1.9} & 79.5 \tiny{$\pm$ 0.8}\\
  \bottomrule
\end{tabular}
  \vspace{0.2cm}
  \caption{The 5-way, 1-shot and 5-shot classification accuracy ($\%$) on miniImageNet and tieredImageNet datasets. ``(tieredPre)'' means the pre-training stage is finished on the tieredImageNet.
  We implement the public code of related methods~\cite{FinnAL17, SnellSZ17, VinyalsBLKW16, SungCVPR2018, chen19closerfewshot} in our framework by which we are able to conduct different meta Operations, i.e. \emph{FT} and \emph{SS}. 
  Note that (1) cosine classifiers have been used in MatchingNets~\cite{SnellSZ17} and Baseline++~\cite{chen19closerfewshot} for few-shot classification; and (2) MAML in this table is not exactly same with original MAML~\cite{FinnAL17}, as it works on deep neural networks and does not update convolutional layers during base-training.
  }
   \label{table_all_heads}
\end{table*}

\begin{table*}[t]
\centering
\small
\begin{tabular}{llccccccccccc}
\toprule 
&& \multicolumn{2}{c}{miniImageNet}
&& \multicolumn{2}{c}{tieredImageNet} && \multicolumn{2}{c}{miniImageNet w/$\mathcal{D}$}
&& \multicolumn{2}{c}{tieredImageNet w/$\mathcal{D}$} \\
\cmidrule{3-4}\cmidrule{6-7} \cmidrule{9-10}\cmidrule{12-13}
&& 1-shot & 5-shot && 1-shot & 5-shot && 1-shot & 5-shot && 1-shot & 5-shot \\
\midrule[0.7pt]
\multicolumn{2}{l}{Masked Soft k-Means~\cite{RenICLR2018_semisupervised}} & 50.4 \tiny{$\pm$ 0.3} & 64.4 \tiny{$\pm$ 0.2} && 52.4 \tiny{$\pm$ 0.4} & 69.9 \tiny{$\pm$ 0.2} && 49.0 \tiny{$\pm$ 0.3} & 63.0 \tiny{$\pm$ 0.1} && 51.4 \tiny{$\pm$ 0.4} & 69.1 \tiny{$\pm$ 0.3} \\
\multicolumn{2}{l}{Masked Soft k-Means \emph{w/} MTL} & 58.2 \tiny{$\pm$ 1.8} & 71.9 \tiny{$\pm$ 0.8} && 65.3 \tiny{$\pm$ 0.9} & 79.8 \tiny{$\pm$ 0.8} && 56.8 \tiny{$\pm$ 1.7} & 71.1 \tiny{$\pm$ 0.8} && 63.6 \tiny{$\pm$ 1.8} & 79.2 \tiny{$\pm$ 0.8} \\
\midrule
\multicolumn{2}{l}{TPN~\cite{LiuICLR2019transductive}} & 52.8 \tiny{$\pm$ 0.3} & 66.4 \tiny{$\pm$ 0.2} && 55.7 \tiny{$\pm$ 0.3} &  71.0 \tiny{$\pm$ 0.2} && 50.4 \tiny{$\pm$ 0.8} &  64.9 \tiny{$\pm$ 0.7} && 53.5 \tiny{$\pm$ 0.9} & 69.9 \tiny{$\pm$ 0.8} \\
\multicolumn{2}{l}{TPN \emph{w/} MTL} & 59.3 \tiny{$\pm$ 1.7} & 71.9 \tiny{$\pm$ 0.8} && 67.4 \tiny{$\pm$ 1.8} & 80.7 \tiny{$\pm$ 0.8} && 58.7 \tiny{$\pm$ 1.7} & 70.6 \tiny{$\pm$ 0.8} && 67.2 \tiny{$\pm$ 1.7} & 80.5 \tiny{$\pm$ 0.9}  \\
\bottomrule[0.7pt]
\end{tabular}
\vspace{0.2cm}
\caption{Semi-supervised 5-way, 1-shot and 5-shot classification accuracy (\%) on miniImageNet and tieredImageNet. ``meta-batch'' and ``ResNet-12 (pre)'' are used. ``w/$\mathcal{D}$'' means additionally including the unlabeled data from $3$ distracting classes ($5$ unlabeled samples per class) that are \textbf{excluded} in the ``5-way'' classes of the task~\cite{LiuICLR2019transductive, RenICLR2018_semisupervised}. 
}
\label{table_semi_withSS}
\end{table*}

The scaling and shifting (\emph{SS}) operations in our proposed MTL approach work on pre-trained convolutional neurons thus are easy to be applied to other CNNs base few-shot learning models.

Table~\ref{table_all_heads} shows the results of implementing \emph{SS} operations on pre-trained deep models and using classical few-shot learning methods, namely ProtoNets~\cite{SnellSZ17}, MatchingNets~\cite{VinyalsBLKW16}, RelationNets~\cite{SungCVPR2018}, 
MAML~\cite{FinnAL17} (with a single FC layer as the base-learner~\cite{SunCVPR2019}),
and MAML~\cite{FinnAL17} (with a cosine distance classifier as the base-learner~\cite{VinyalsBLKW16, chen19closerfewshot}), under fair data split settings. 
In the original versions of those methods, \emph{FT} is the meta-level operation and 4CONV is the uniform architecture. For comparison, we, therefore, implement the results of using \emph{FT} on deeper networks -- ResNet-12 and ResNet-18.
In Teble~\ref{table_all_heads}, we have three column results. ``miniImageNet(tieredPre)'' denotes that the model is pre-trained on tieredImageNet and its weights are then meta-transferred to the learning of few-shot models on miniImageNet tasks.
We can see that in all settings, (1) the best performance is always achieved by our proposed \emph{SS} operations, e.g., MTL (FC, ResNet-12) outperforms MAML (FC, ResNet-12) by $3.6\%$ and $8.1\%$ on tieredImageNet 1-shot and 5-shot, respectively.; and (2) each classical method using \emph{SS} gets consistent improvements its original \emph{FT} version,
e.g., RelationNets~\cite{SungCVPR2018} (ResNet-18) gains $6.1\%$ and $6.7\%$ on miniImageNet 1-shot and 5-shot, respectively.

In addition, we verify the generalization ability of our MTL to the state-of-the-art semi-supervised few-shot learning (SSFSL) methods~\cite{LiuICLR2019transductive, RenICLR2018_semisupervised}.
To this end, we conduct experiments in the uniform SSFSL settings with $5$ unlabeled samples per class added to the episode training set, following~\cite{LiuICLR2019transductive, RenICLR2018_semisupervised}. Our results on the miniImageNet and tieredImageNet datasets are presented in Table~\ref{table_semi_withSS}. Note that ``w/$\mathcal{D}$'' indicates the more challenging setting with $3$ distracting classes added to the unlabeled data of the task.
In each block of Table~\ref{table_semi_withSS}, we can clearly see that the methods ``\emph{w}/ MTL'' obtain consistent performance improvements over the original methods in both normal and challenging SSFSL settings by quit large margins, e.g. $13.7\%$ on tieredImageNet w/$\mathcal{D}$ 1-shot. This validates the effectiveness of our method for tackling SSFSL problems.

\subsection{Ablation study}
\label{sec_ablation_study}

\begin{table*}[t]
\centering
% \vspace{-0.4cm}
\small
\begin{tabular}{lcccccccccccc}
\toprule 
& \multicolumn{2}{c}{\multirow{2}{*}{miniImageNet}} & &
\multicolumn{2}{c}{miniImageNet}
&& \multicolumn{2}{c}{\multirow{2}{*}{FC100}} & & \multicolumn{2}{c}{FC100}  \\
&&&&\multicolumn{2}{c}{(tieredPre)}&&&&&\multicolumn{2}{c}{(tieredPre)}\\
\cmidrule{2-3}\cmidrule{5-6} \cmidrule{8-9} \cmidrule{11-12} 
& 1-shot & 5-shot & & 1-shot & 5-shot & & 1-shot & 5-shot & & 1-shot & 5-shot \\
\midrule[0.7pt]
\emph{update} $[\Theta; \theta]$ & 45.3 \tiny{$\pm$ 1.9} & 64.6 \tiny{$\pm$ 0.9} && 54.4 \tiny{$\pm$ 1.8} & 73.7 \tiny{$\pm$ 0.8} && 38.4 \tiny{$\pm$ 1.8} & 52.6 \tiny{$\pm$ 0.9} && 37.6 \tiny{$\pm$ 1.9}&52.7 \tiny{$\pm$ 0.9} \\
\emph{update} $\theta$ & 50.0 \tiny{$\pm$ 1.8} & 66.7 \tiny{$\pm$ 0.9} && 51.3 \tiny{$\pm$ 1.8} & 70.3 \tiny{$\pm$ 0.8} && 39.3 \tiny{$\pm$ 1.9} & 51.8 \tiny{$\pm$ 0.9} && 38.3 \tiny{$\pm$ 1.8} & 52.9 \tiny{$\pm$ 1.0} \\
\midrule
% \midrule[0.7pt]
\emph{FT} $\theta$  & 55.9 \tiny{$\pm$ 1.9} & 71.4 \tiny{$\pm$ 0.9} && 61.6 \tiny{$\pm$ 1.8} & 73.5 \tiny{$\pm$ 0.9} && 41.6 \tiny{$\pm$ 1.9} & 54.9 \tiny{$\pm$ 1.0} && 40.4 \tiny{$\pm$ 1.9} & 54.7 \tiny{$\pm$ 0.9} \\
\emph{FT} $[\Theta4; \theta]$ & 57.2 \tiny{$\pm$ 1.8} & 71.6 \tiny{$\pm$ 0.8} && 62.3 \tiny{$\pm$ 1.8} & 73.9 \tiny{$\pm$ 0.9} && 40.9 \tiny{$\pm$ 1.8} & 54.3 \tiny{$\pm$ 1.0} && 41.2 \tiny{$\pm$ 1.8} &53.6 \tiny{$\pm$ 1.0}  \\
\emph{FT} $[\Theta3,\Theta4; \theta]$ & 58.1 \tiny{$\pm$ 1.8} & 70.9 \tiny{$\pm$ 0.8} && 63.0 \tiny{$\pm$ 1.8} & 74.8 \tiny{$\pm$ 0.9} && 41.5 \tiny{$\pm$ 1.8} & 53.7 \tiny{$\pm$ 0.9} && 40.7 \tiny{$\pm$ 1.9} &53.8 \tiny{$\pm$ 0.9} \\
\emph{FT} $[\Theta; \theta]$  & 58.3 \tiny{$\pm$ 1.8} & 71.6 \tiny{$\pm$ 0.8} && 63.2 \tiny{$\pm$ 1.9} & 75.7 \tiny{$\pm$ 0.8}&& 41.6 \tiny{$\pm$ 1.9} & 54.4 \tiny{$\pm$ 1.0} && 41.1 \tiny{$\pm$ 1.9}& 54.5 \tiny{$\pm$ 0.9} \\
\midrule
\emph{SS} $[\Theta4; \theta]$ & 59.2 \tiny{$\pm$ 1.8} & 73.1 \tiny{$\pm$ 0.9} && 64.0 \tiny{$\pm$ 1.8} & 76.9 \tiny{$\pm$ 0.8} && 42.4 \tiny{$\pm$ 1.9} & 55.1 \tiny{$\pm$ 1.0} &&42.7 \tiny{$\pm$ 1.9} &55.9 \tiny{$\pm$ 1.0} \\
\emph{SS} $[\Theta3,\Theta4; \theta]$ & 59.4 \tiny{$\pm$ 1.8} & 73.4 \tiny {$\pm$ 0.8} && 64.5 \tiny{$\pm$ 1.8} & 77.2 \tiny{$\pm$ 0.8} && 42.5 \tiny{$\pm$ 1.9} & 54.5 \tiny{$\pm$ 1.0} && 43.4 \tiny{$\pm$ 1.8}&56.4 \tiny{$\pm$ 1.0} \\
% \midrule[0.7pt]
\emph{SS $[\Theta; \theta]$}\textbf{(Ours)} & \textbf{60.6} \tiny{$\pm$ 1.8} & \textbf{74.3} \tiny{$\pm$ 0.8} && \textbf{65.7} \tiny{$\pm$ 1.8} & \textbf{78.4} \tiny{$\pm$ 0.9} && \textbf{43.6} \tiny{$\pm$ 1.9} & \textbf{55.4} \tiny{$\pm$ 1.0} && \textbf{43.5} \tiny{$\pm$ 1.9} & \textbf{57.1} \tiny{$\pm$ 1.0}  \\
\bottomrule[0.7pt]
\end{tabular}
\vspace{0.1cm}
\caption{The 5-way, 1-shot and 5-shot classification accuracy (\%) using ablative models, on two datasets. ``meta-batch'' and ``ResNet-12 (pre)'' are used. ``(tieredPre)'' means the pre-training stage is finished on the tieredImageNet.}
\label{table_ablation}
\end{table*}
\begin{figure*}
\newcommand{\newincludegraphics}[1]{\includegraphics[height=1.25in]{#1}}
\centering
\subfigure[\emph{FT}, ResNet-12, miniImageNet]{
\newincludegraphics{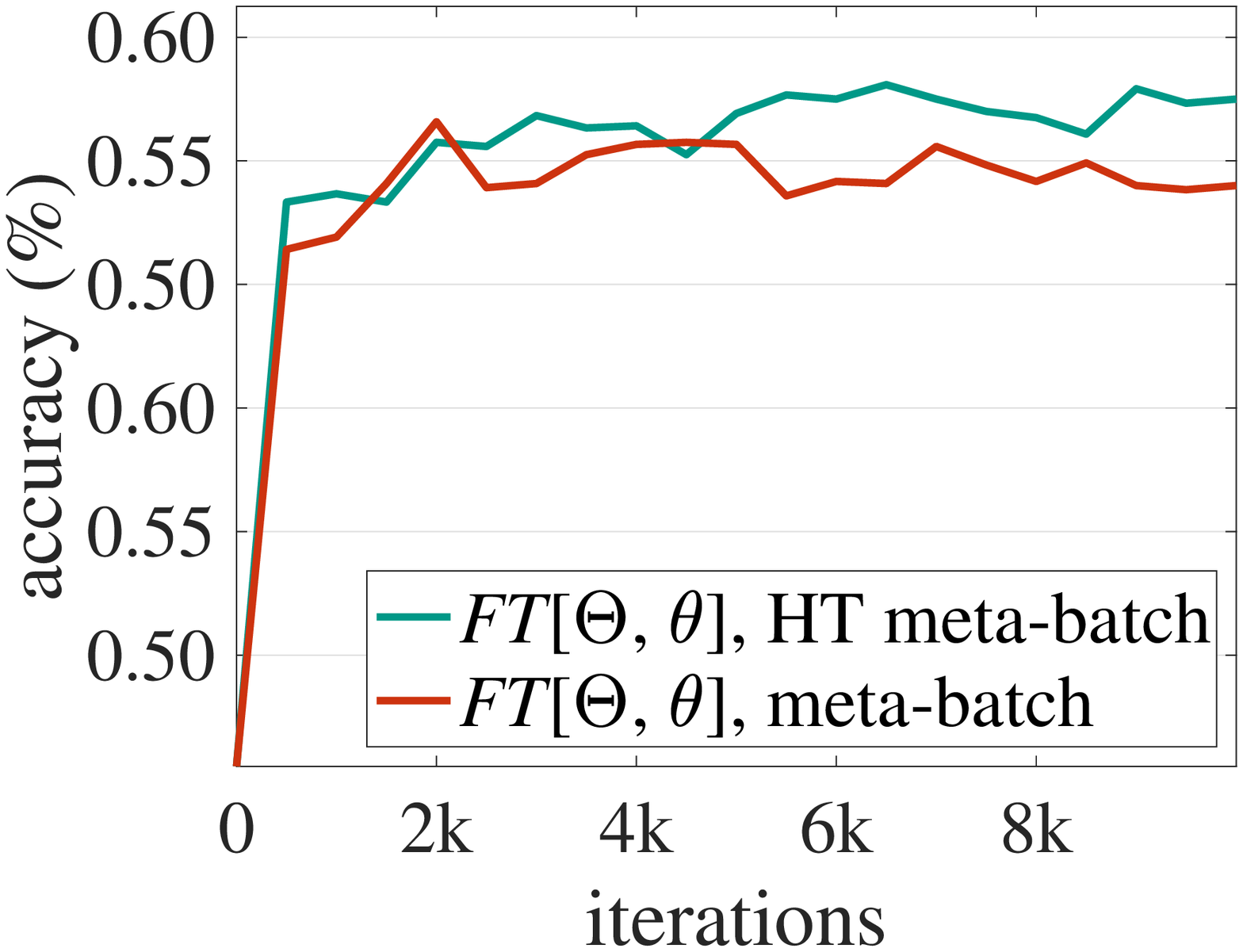}
}
\subfigure[\emph{SS}, ResNet-12, miniImageNet]{
\newincludegraphics{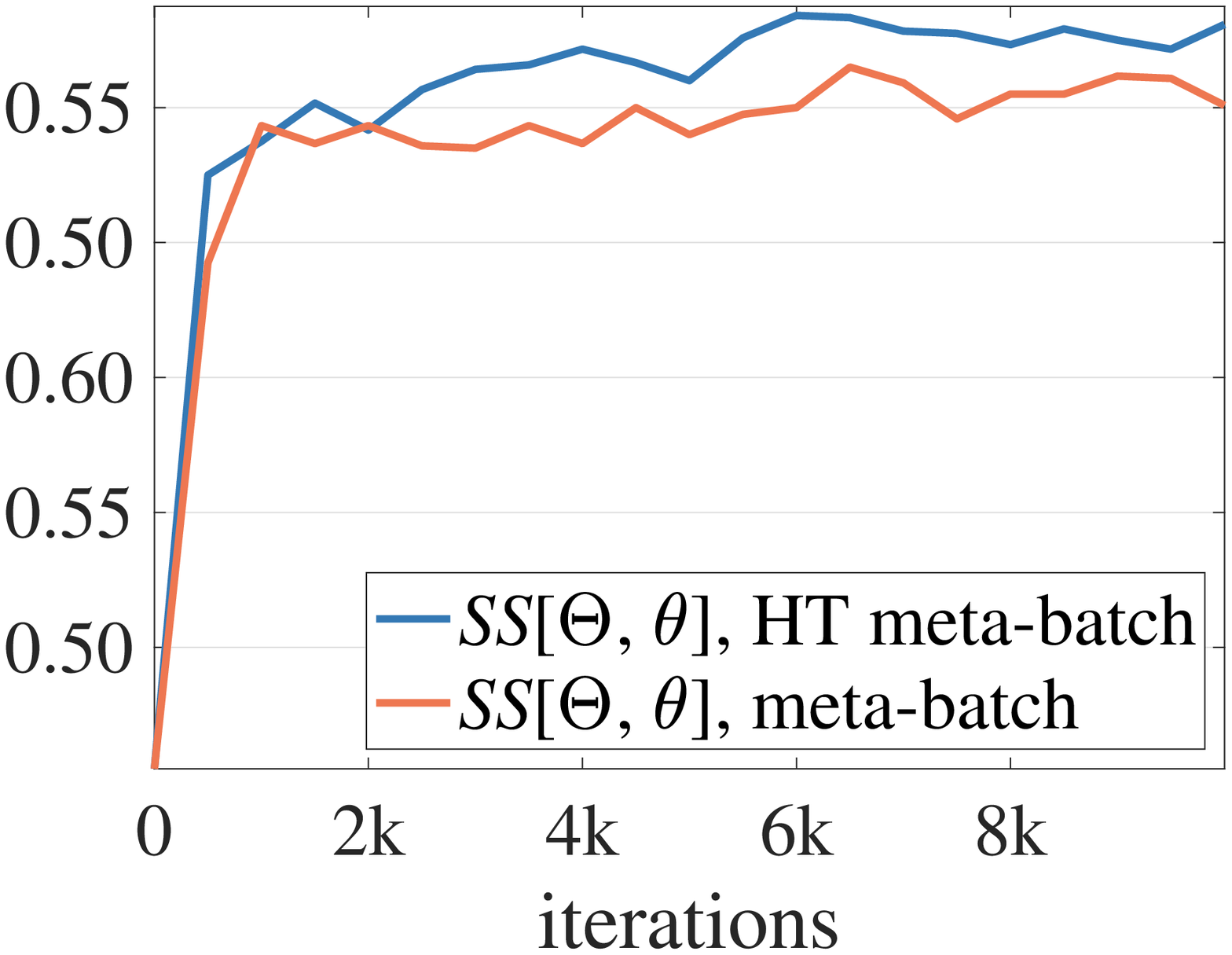}
}
\subfigure[\emph{SS}, ResNet-18, miniImageNet]{
\newincludegraphics{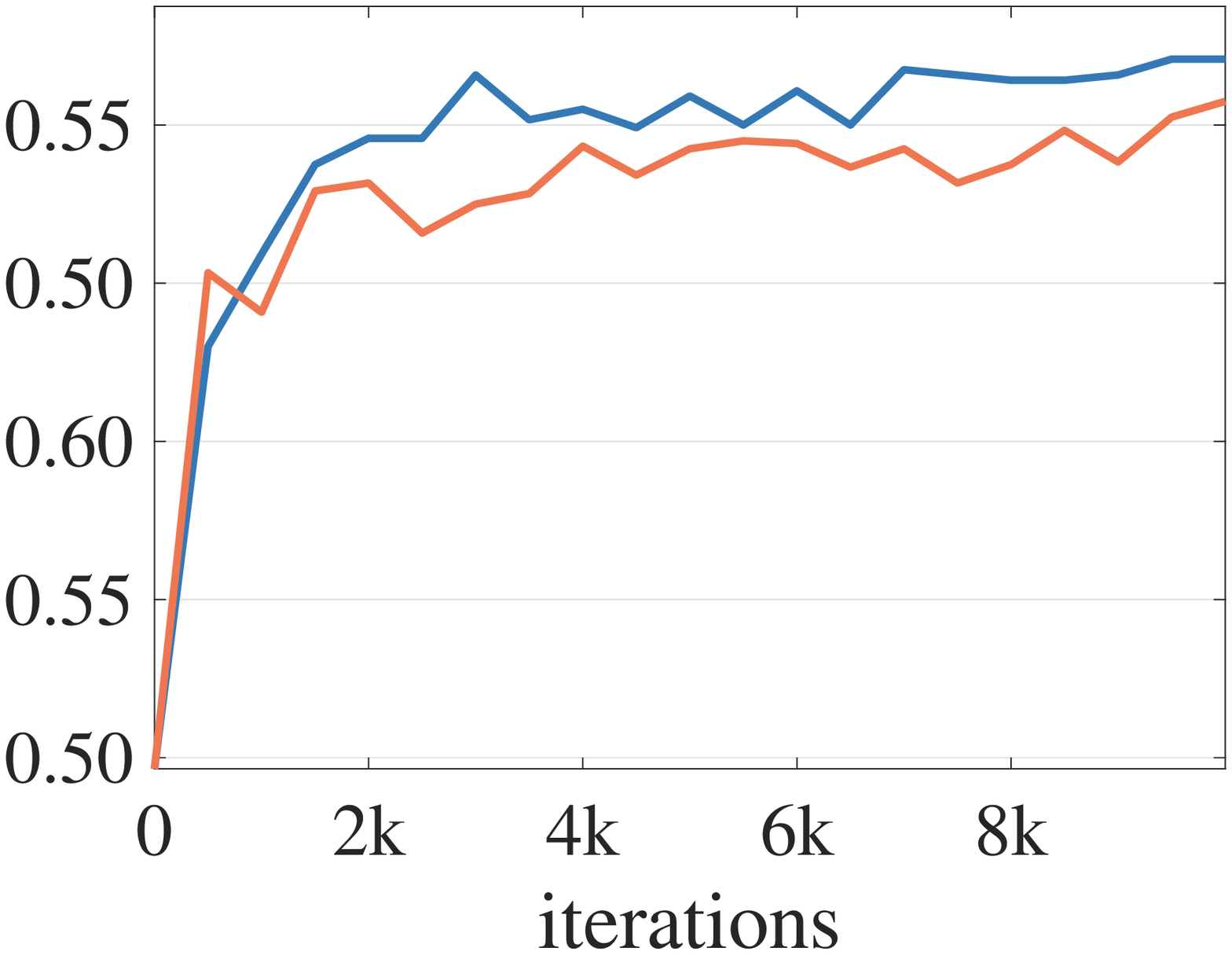}
}
\subfigure[\emph{SS}, ResNet-25, miniImageNet]{
\newincludegraphics{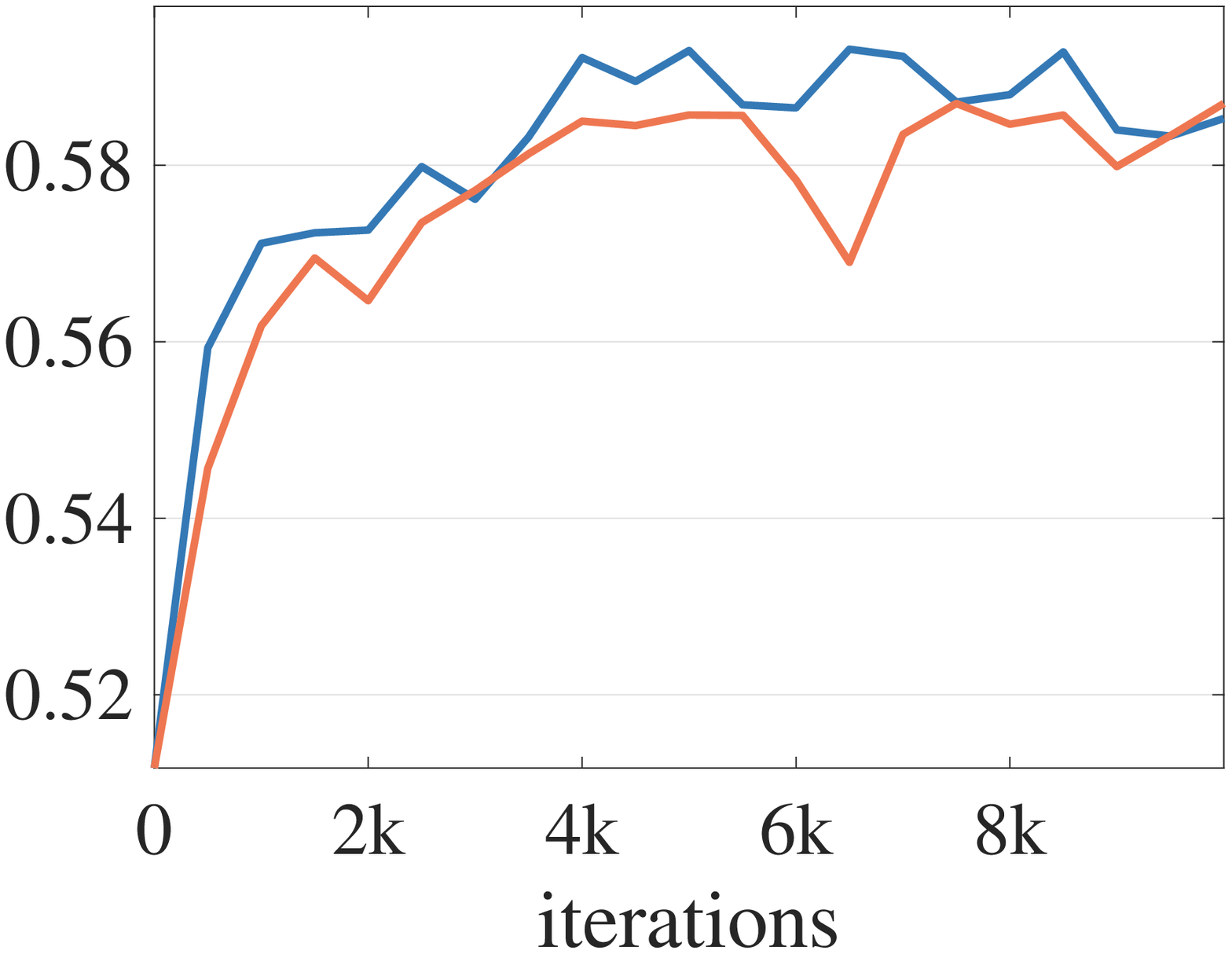}
}
\\
\subfigure[\emph{FT}, ResNet-12, tieredImageNet]{
\newincludegraphics{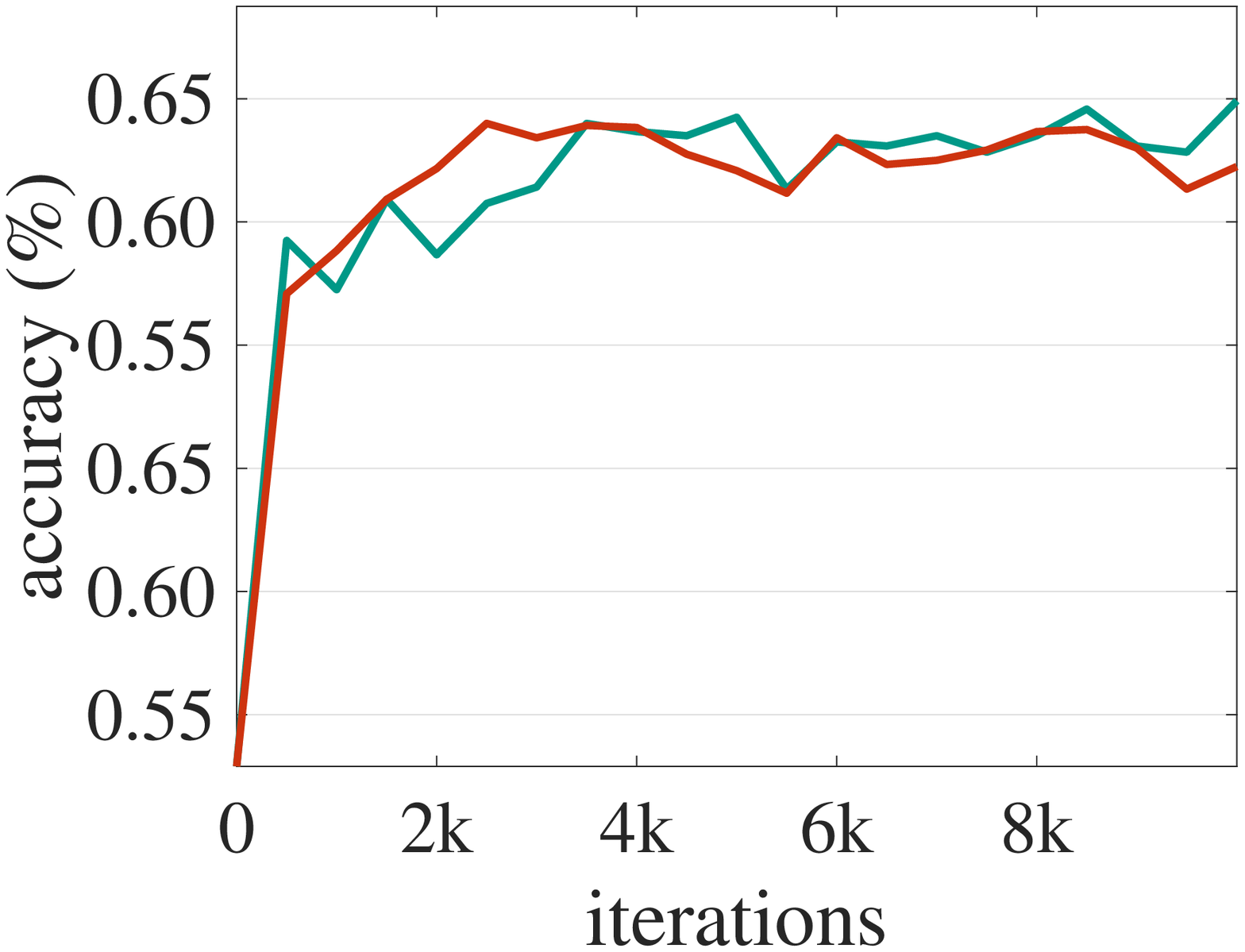}
}
\subfigure[\emph{SS}, ResNet-12, tieredImageNet]{
\newincludegraphics{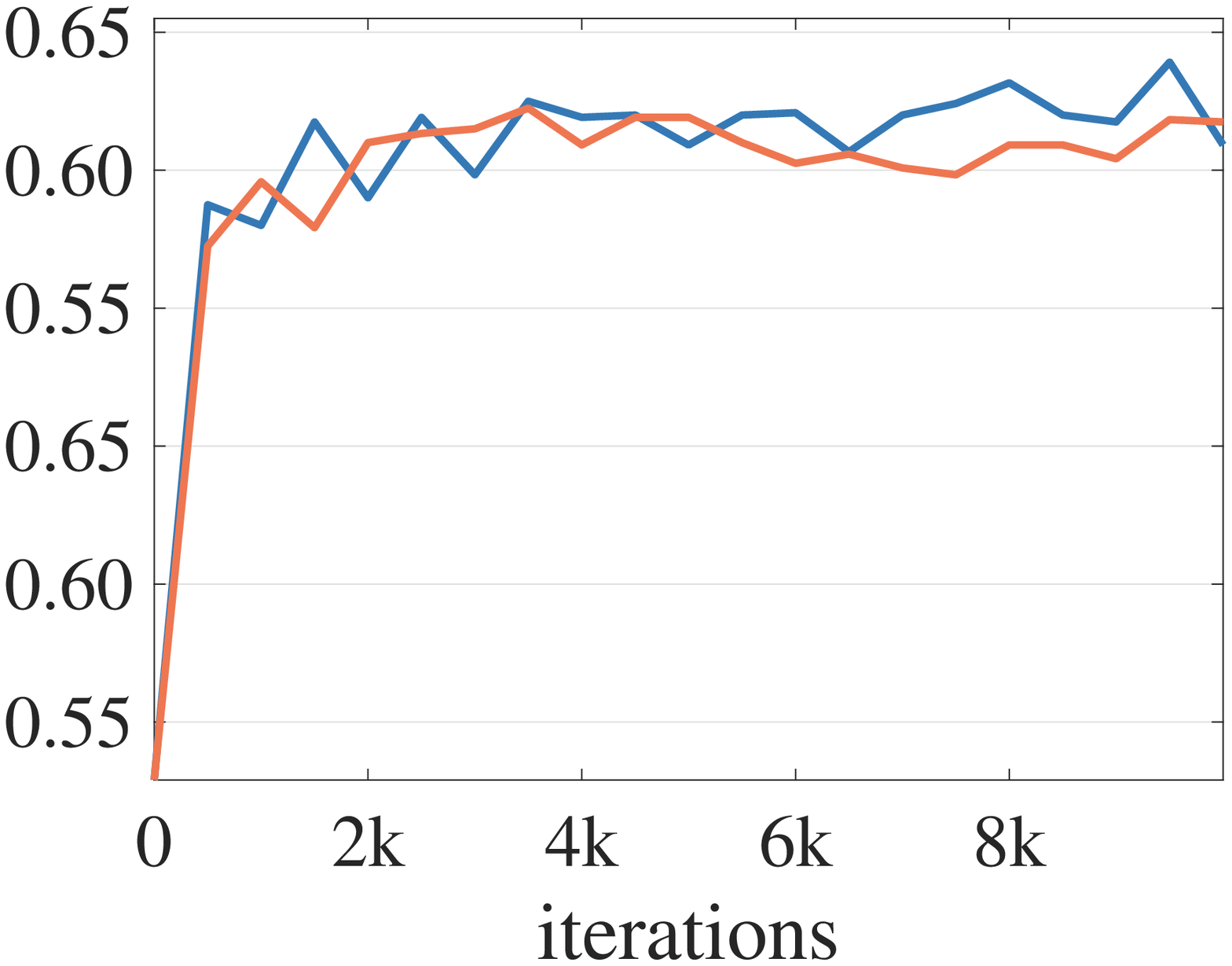}
}
\subfigure[\emph{SS}, ResNet-18, tieredImageNet]{
\newincludegraphics{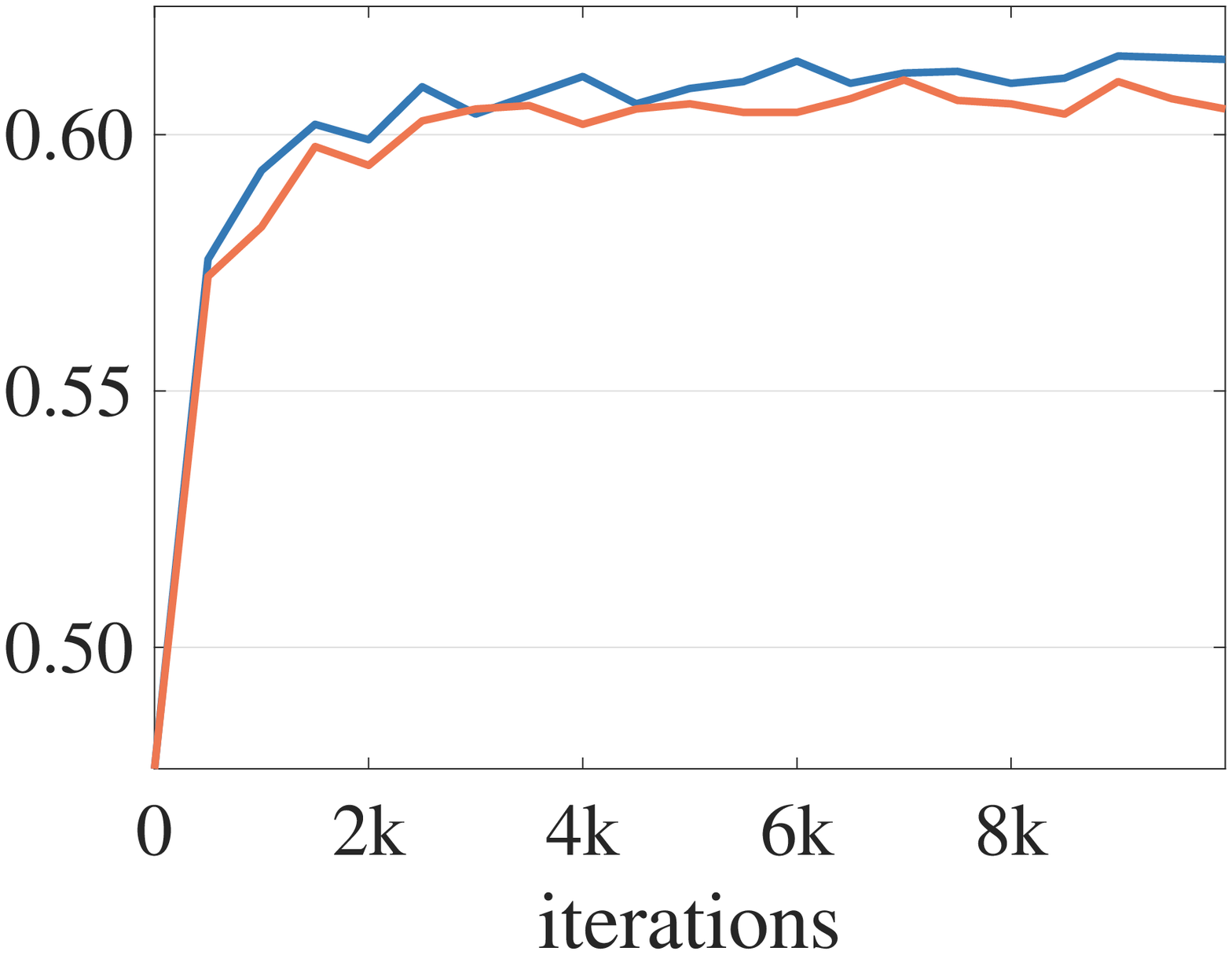}
}
\subfigure[\emph{SS}, ResNet-25, tieredImageNet]{
\newincludegraphics{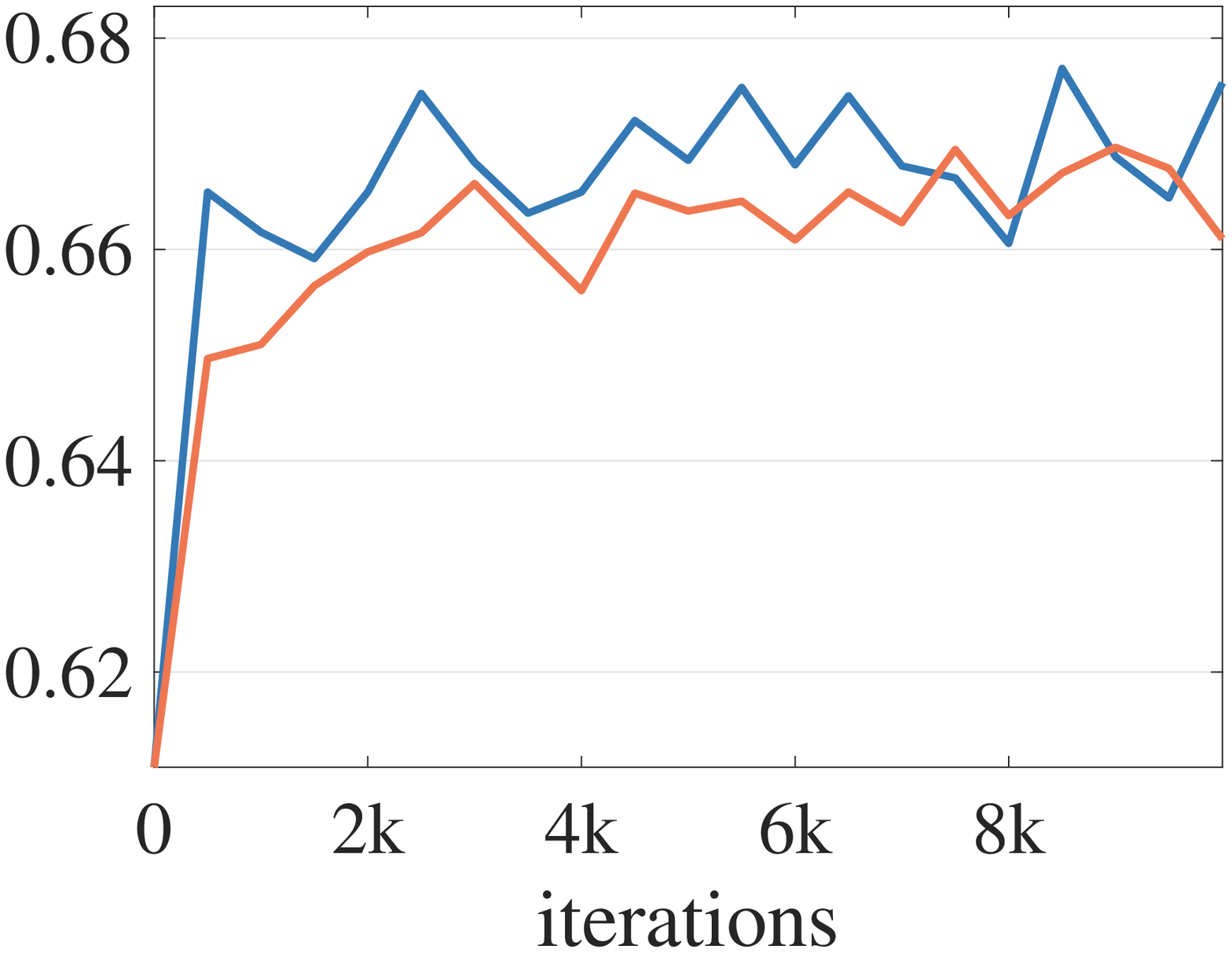}
}
\\
\subfigure[\emph{FT}, ResNet-12, FC100]{
\newincludegraphics{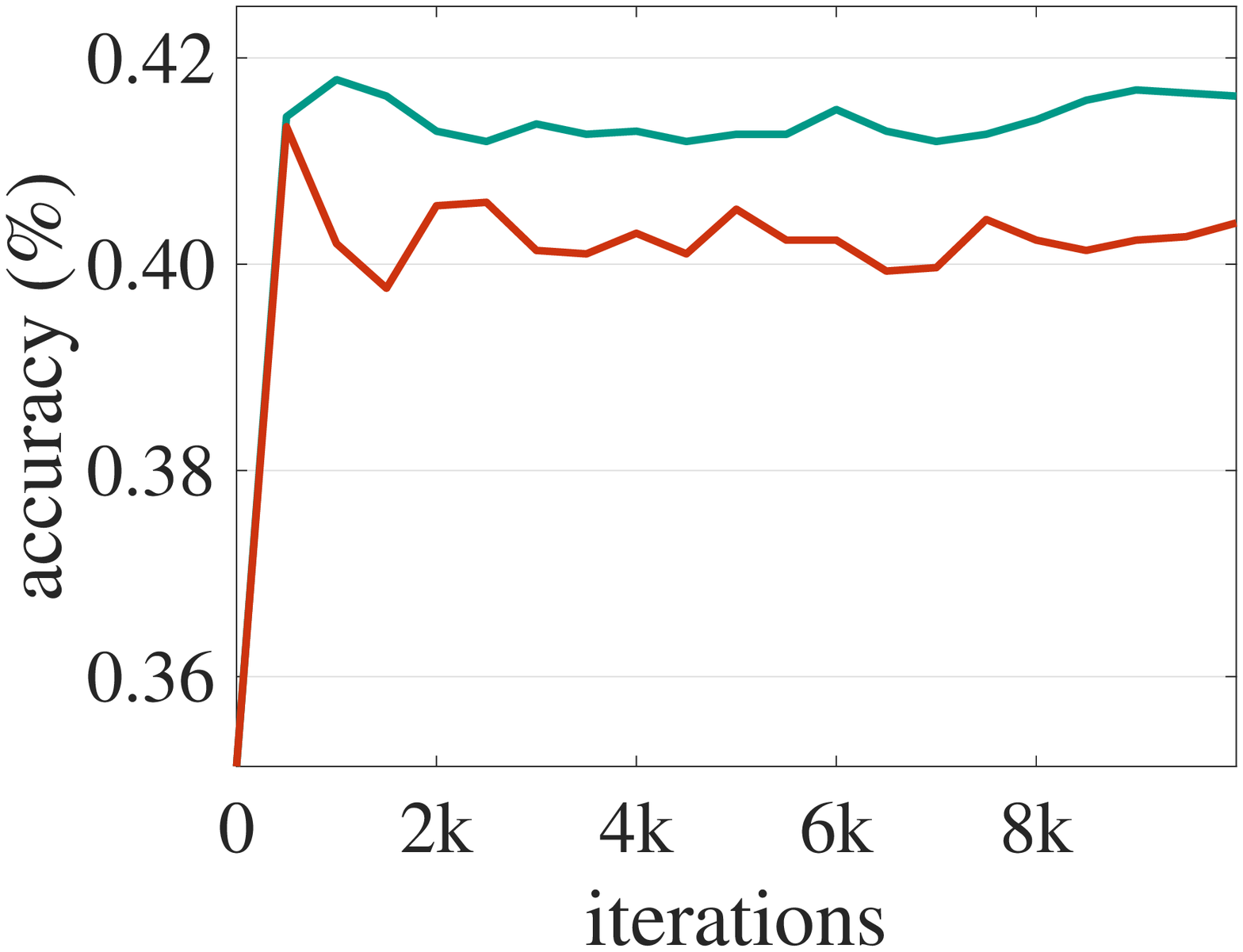}
}
\subfigure[\emph{SS}, ResNet-12, FC100]{
\newincludegraphics{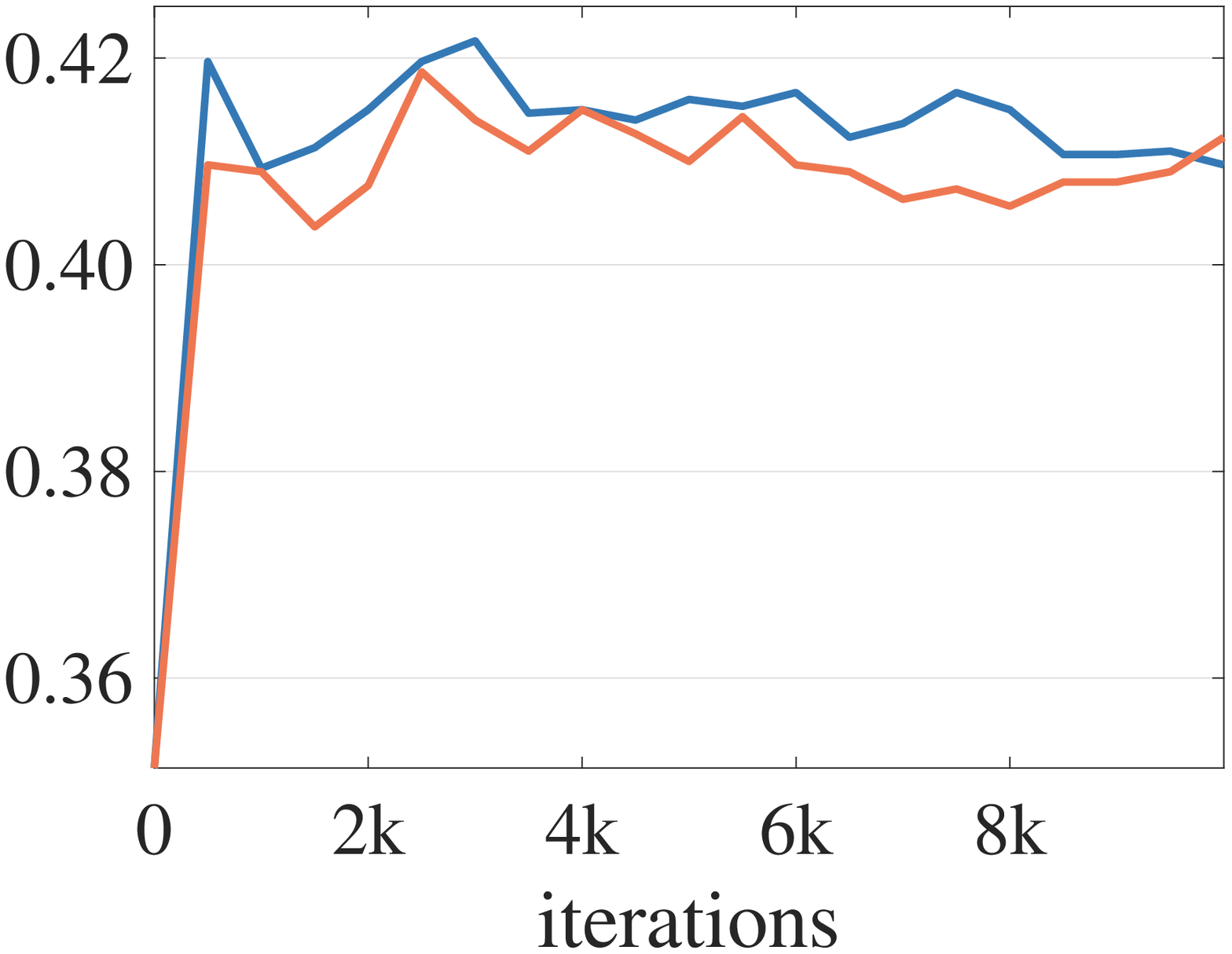}
}
\subfigure[\emph{SS}, ResNet-18, FC100]{
\newincludegraphics{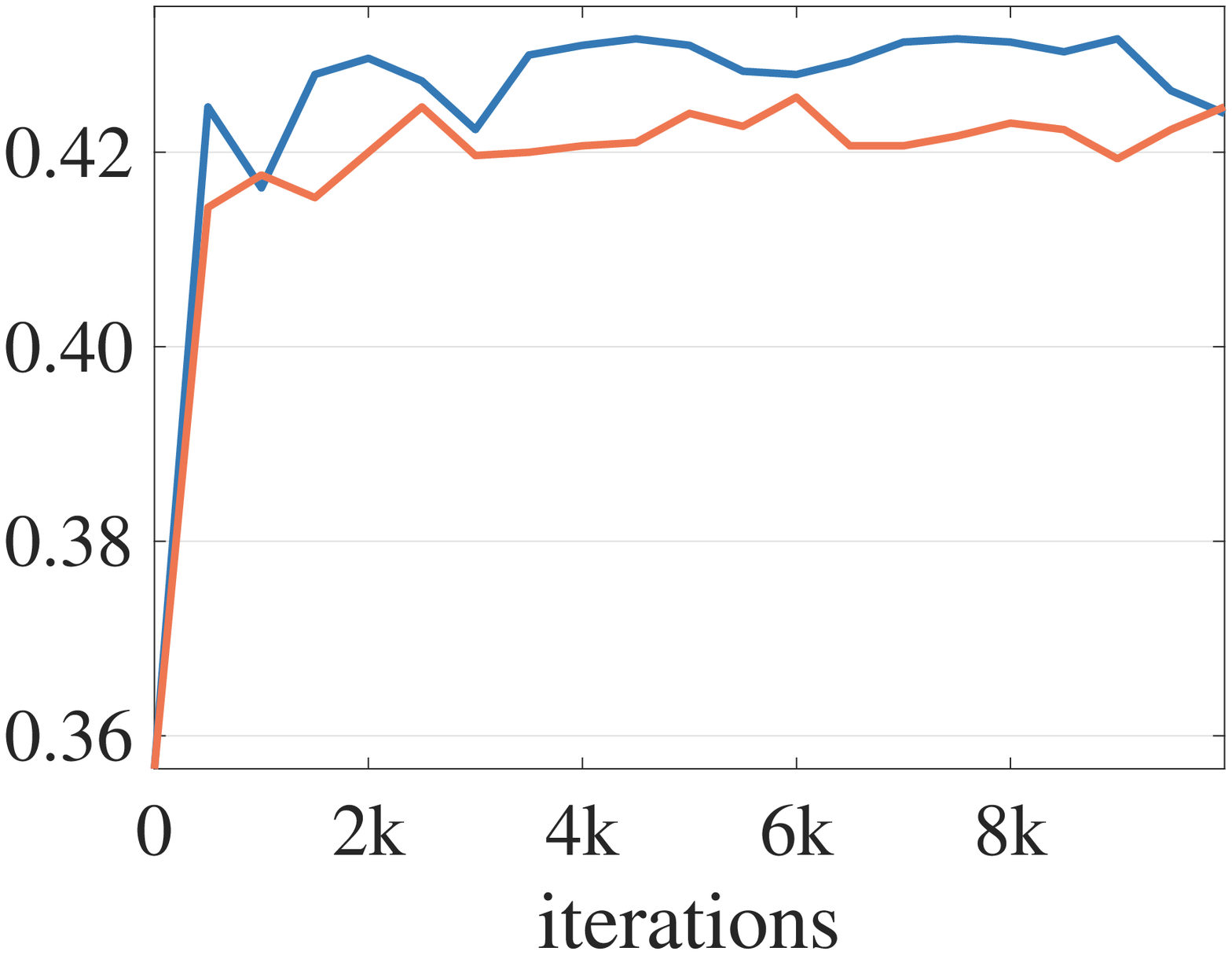}
}
\subfigure[\emph{SS}, ResNet-25, FC100]{
\newincludegraphics{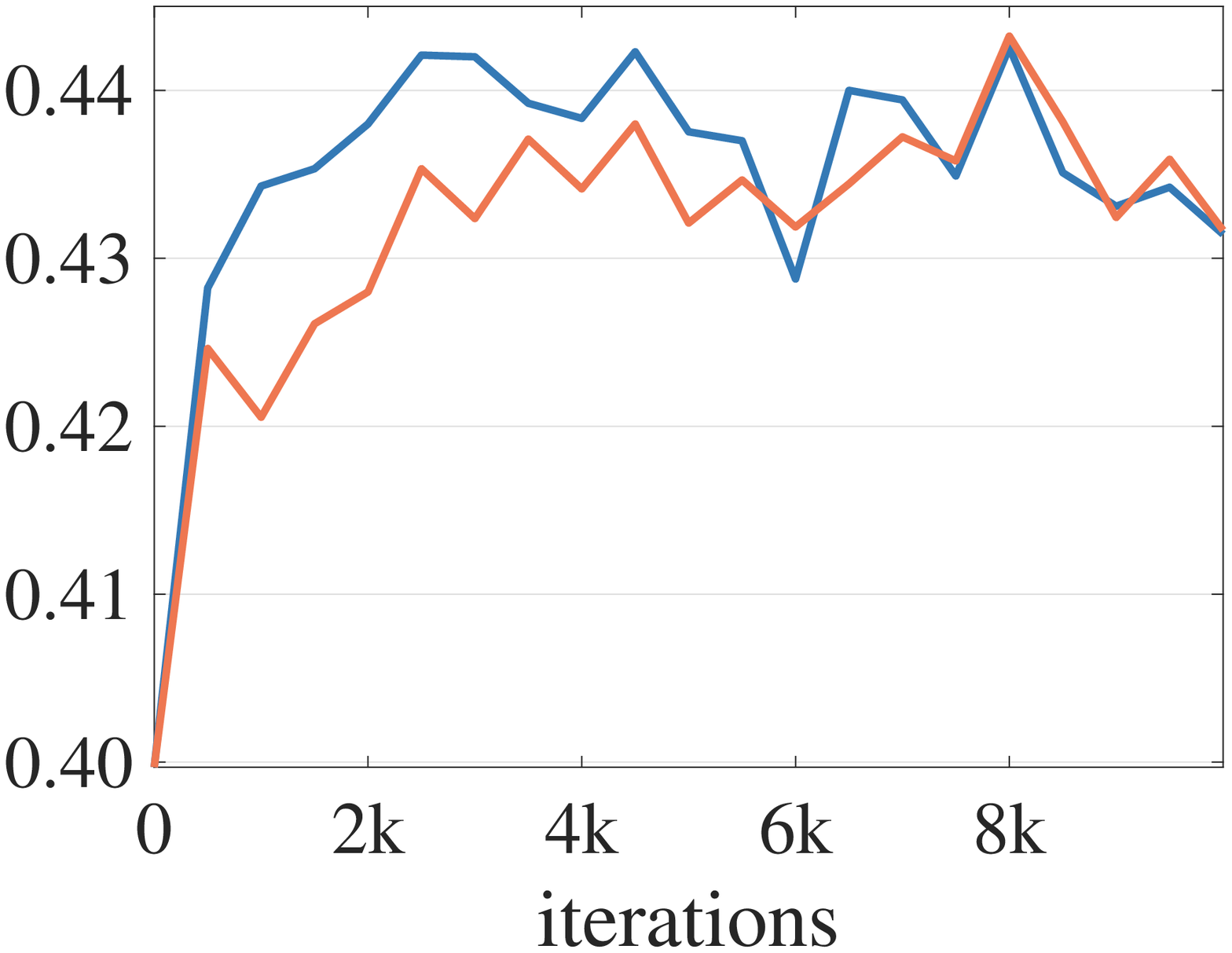}
}
\vspace{0.2cm}
\caption{The 5-way, 1-shot meta-validation accuracy ($\%$) of using \emph{FT} (deep pre-trained network based MAML~\cite{FinnAL17}) and \emph{SS} (our MTL) with original meta-batch~\cite{FinnAL17} and our proposed HT meta-batch.}
\label{fig_ht_meta_batch}
\end{figure*}

Table~\ref{table_ablation} shows the extensive results of MTL-related ablative methods, on the miniImageNet and FC100 datasets.
Figure~\ref{fig_ht_meta_batch} demonstrates the performance gap between \emph{w/} and \emph{w/o} HT meta-batch in terms of accuracy and converging speed.
Extensive comparisons are conducted on three datasets, three network architectures and two meta-level operations.

%%%%%%%%%%%%%%%%%%%%

\myparagraph{MTL \emph{vs}. No meta-learning.}
Table~\ref{table_ablation} shows the results of \emph{No meta-learning} methods on the top block.
Compared to these, our approach achieves significantly better performance,
e.g. the largest margins on miniImageNet are $10.6\%$ for 1-shot and $7.6\%$ for 5-shot.
This validates the effectiveness of our meta-learning method for tackling few-shot learning problems.
Between two \emph{No meta-learning} methods, we can see that updating both feature extractor $\Theta$ and classifier $\theta$ is inferior to updating $\theta$ only ($\Theta$ is pre-trained), e.g. around $5\%$ reduction on miniImageNet 1-shot.
One reason is that in few-shot settings, there are too many parameters to optimize with little data.
This supports our motivation to learn only $\theta$ during base-learning.

%%%%%%%%%%%%%%%%%%%%%%%%%%%%%%%%%%%%%%%%%%%%%%%%%%%%%%%%%%%%%
\myparagraph{\emph{SS} $[\Theta; \theta]$ works better than light-weight \emph{FT} variants.} 
Table~\ref{table_ablation} shows that our approach with \emph{SS} $[\Theta; \theta]$ achieves the best performances for all few-shot settings. 
In principle, \emph{SS} meta-learns a smaller set of transferring parameters on $[\Theta; \theta]$ than \emph{FT}.
People may argue that \emph{FT} is weaker because it learns a larger set of initialization parameters, whose quantity is equal to the size of $[\Theta; \theta]$, causing the overfit to the few-shot data.  
In the middle block of Table~\ref{table_ablation},
we show the ablation study of freezing low-level pre-trained layers and meta-learn only the high-level layers (e.g. the $4$-th residual block $\Theta4$ of ResNet-12) by the \emph{FT} operations.
It is obvious that they all yield inferior performances than using our \emph{SS}. 
An additional observation is that \emph{SS}* performs consistently better than \emph{FT}*.

\myparagraph{Accuracy gain by HT meta-batch.}
HT meta-batch is basically a curriculum learning scheme, and can be generalized to the models with different network architectures.  
Comparing the whole model, i.e. \emph{SS} $[\Theta; \theta]$ - HT meta-batch - ResNet-12(pre) (in Table~\ref{table_sota_all}), with the \emph{SS} $[\Theta; \theta]$ in Table~\ref{table_ablation}, we can observe that HT meta-batch brings an accuracy boost in a range of $0.6\% \sim 2.2\%$. It is interesting to observe that on the more challenging FC100 dataset it brings an average gain of $1.85\%$, more than twice of the $0.9\%$ on miniImageNet. In addition, Figure~\ref{fig_ht_meta_batch} shows the validation results of the models obtained on different meta-training iterations. It demonstrates that our HT meta-batch consistently achieves higher performances than the conventional meta-batch~\cite{FinnAL17}. This ability is also shown in the first column for \emph{FT} based methods (pre-trained ResNet-12 based MAML~\cite{FinnAL17}).

\myparagraph{Speed of convergence of MTL with HT meta-batch.} 
MAML~\cite{FinnAL17} used $240k$ tasks to achieve the best performance on the miniImageNet.
Impressively, our HT meta-batch methods used only around $8k$ tasks to converge, see Figure~\ref{fig_ht_meta_batch} (b)-(d) (note that each iteration contains $2$ tasks). 
This advantage is more obvious for FC100 on which HT meta-batch methods need at most $6k$ tasks, Figure~\ref{fig_ht_meta_batch}(g)-(i).
We attest this to three reasons. First, our methods start from the pre-trained deep neural networks. 
Second, \emph{SS} needs to learn only $< \tfrac{2}{9}$ parameters of the number of \emph{FT} parameters.
Third, HT meta-batch is a hard negative mining step and brings accelerations by learning the challenging tasks~\cite{ShrivastavaGG16}.

%

%%%%%%%%%%%%%%%%%%%%%%%%%%%%%%%%%%%%%%

\section{Conclusions}

In this paper, we show that our novel MTL model trained with HT meta-batch learning curriculum achieves the top performance for tackling few-shot learning problems.
The key operations of MTL on pre-trained DNN neurons proved to be highly efficient for adapting learning experience to the unseen task.
The superiority was particularly achieved in the extreme 1-shot cases on three challenging benchmarks -- miniImageNet, tieredImageNet and FC100. The generalization ability of our method is validated by implementing MTL on the classical supervised few-shot models as well as the state-of-the-art semi-supervised few-shot models.
The consistent improvements by MTL prove that large-scale pre-trained deep networks can offer a good ``knowledge base'' to conduct efficient few-shot learning on.
In terms of learning scheme, HT meta-batch showed consistently good performance for the ablative models.
On the more challenging FC100 benchmark, it showed to be particularly helpful for boosting convergence speed.
This design is independent of any specific model or architecture and can be generalized well whenever the hardness of task is easy to evaluate in online iterations.

\section*{Acknowledgments}
This research is part of NExT research which is supported by the National Research Foundation, Prime Minister's Office, Singapore under its IRC@SG Funding Initiative.
It is also partially supported by German Research Foundation (DFG CRC 1223), and National Natural Science Foundation of China (61772359).

% \appendices

% % you can choose not to have a title for an appendix
% % if you want by leaving the argument blank
% \section{}

% % use section* for acknowledgment
% \ifCLASSOPTIONcompsoc
%   % The Computer Society usually uses the plural form
%   \section*{Acknowledgments}
% \else
%   % regular IEEE prefers the singular form
% \fi


\ifCLASSOPTIONcaptionsoff
  \newpage
\fi

\bibliography{egbib.bib}{}

\begin{thebibliography}{10}
\providecommand{\url}[1]{#1}
\csname url@samestyle\endcsname
\providecommand{\newblock}{\relax}
\providecommand{\bibinfo}[2]{#2}
\providecommand{\BIBentrySTDinterwordspacing}{\spaceskip=0pt\relax}
\providecommand{\BIBentryALTinterwordstretchfactor}{4}
\providecommand{\BIBentryALTinterwordspacing}{\spaceskip=\fontdimen2\font plus
\BIBentryALTinterwordstretchfactor\fontdimen3\font minus
  \fontdimen4\font\relax}
\providecommand{\BIBforeignlanguage}[2]{{%
\expandafter\ifx\csname l@#1\endcsname\relax
\typeout{** WARNING: IEEEtran.bst: No hyphenation pattern has been}%
\typeout{** loaded for the language `#1'. Using the pattern for}%
\typeout{** the default language instead.}%
\else
\language=\csname l@#1\endcsname
\fi
#2}}
\providecommand{\BIBdecl}{\relax}
\BIBdecl

\bibitem{SunCVPR2019}
Q.~Sun, Y.~Liu, T.-S. Chua, and B.~Schiele, ``Meta-transfer learning for
  few-shot learning,'' in \emph{CVPR}, 2019.

\bibitem{RenICLR2018_semisupervised}
M.~Ren, E.~Triantafillou, S.~Ravi, J.~Snell, K.~Swersky, J.~B. Tenenbaum,
  H.~Larochelle, and R.~S. Zemel, ``Meta-learning for semi-supervised few-shot
  classification,'' in \emph{ICLR}, 2018.

\bibitem{Lecun2015}
L.~Yann, B.~Yoshua, and H.~Geoffrey, ``Deep learning,'' \emph{Nature}, vol.
  521(7553), p. 436, 2015.

\bibitem{HeZRS16}
K.~He, X.~Zhang, S.~Ren, and J.~Sun, ``Deep residual learning for image
  recognition,'' in \emph{CVPR}, 2016.

\bibitem{ShelhamerLD17}
E.~Shelhamer, J.~Long, and T.~Darrell, ``Fully convolutional networks for
  semantic segmentation,'' \emph{{IEEE} Trans. Pattern Anal. Mach. Intell.},
  vol.~39, no.~4, pp. 640--651, 2017.

\bibitem{FeiFeiFP06}
F.~Li, R.~Fergus, and P.~Perona, ``One-shot learning of object categories,''
  \emph{{IEEE} Trans. Pattern Anal. Mach. Intell.}, vol.~28, no.~4, pp.
  594--611, 2006.

\bibitem{FinnAL17}
C.~Finn, P.~Abbeel, and S.~Levine, ``Model-agnostic meta-learning for fast
  adaptation of deep networks,'' in \emph{ICML}, 2017.

\bibitem{ClevertUH15}
D.~Clevert, T.~Unterthiner, and S.~Hochreiter, ``Fast and accurate deep network
  learning by exponential linear units (elus),'' in \emph{ICLR}, 2016.

\bibitem{PanTKY11}
S.~J. Pan, I.~W. Tsang, J.~T. Kwok, and Q.~Yang, ``Domain adaptation via
  transfer component analysis,'' \emph{{IEEE} Trans. Neural Networks}, vol.~22,
  no.~2, pp. 199--210, 2011.

\bibitem{KhorevaBIBS17}
A.~Khoreva, R.~Benenson, E.~Ilg, T.~Brox, and B.~Schiele, ``Lucid data dreaming
  for object tracking,'' \emph{arXiv}, vol. 1703.09554, 2017.

\bibitem{Mehrotra2017}
A.~Mehrotra and A.~Dukkipati, ``Generative adversarial residual pairwise
  networks for one shot learning,'' \emph{arXiv}, vol. 1703.08033, 2017.

\bibitem{SchwartzNIPS18}
E.~Schwartz, L.~Karlinsky, J.~Shtok, S.~Harary, M.~Marder, R.~S. Feris,
  A.~Kumar, R.~Giryes, and A.~M. Bronstein, ``Delta-encoder: an effective
  sample synthesis method for few-shot object recognition,'' in \emph{NeurIPS},
  2018.

\bibitem{WangCVPR2018}
Y.~Wang, R.~B. Girshick, M.~Hebert, and B.~Hariharan, ``Low-shot learning from
  imaginary data,'' in \emph{CVPR}, 2018.

\bibitem{XianCVPR2019a}
Y.~Xian, S.~Sharma, B.~Schiele, and Z.~Akata, ``{f-VAEGAN-D2}: {A} feature
  generating framework for any-shot learning,'' in \emph{CVPR}, 2019.

\bibitem{BartunovV18}
S.~Bartunov and D.~P. Vetrov, ``Few-shot generative modelling with generative
  matching networks,'' in \emph{AISTATS}, 2018.

\bibitem{Bengio92}
S.~Bengio, Y.~Bengio, J.~Cloutier, and J.~Gecsei, ``On the optimization of a
  synaptic learning rule,'' in \emph{Optimality in Artificial and Biological
  Neural Networks}.\hskip 1em plus 0.5em minus 0.4em\relax Univ. of Texas,
  1992, pp. 6--8.

\bibitem{Naik92}
D.~K. Naik and R.~Mammone, ``Meta-neural networks that learn by learning,'' in
  \emph{IJCNN}, 1992.

\bibitem{Thrun1998}
S.~Thrun and L.~Pratt, ``Learning to learn: Introduction and overview,'' in
  \emph{Learning to learn}.\hskip 1em plus 0.5em minus 0.4em\relax Springer,
  1998, pp. 3--17.

\bibitem{FinnNIPS2018}
C.~Finn, K.~Xu, and S.~Levine, ``Probabilistic model-agnostic meta-learning,''
  in \emph{NeurIPS}, 2018.

\bibitem{GrantICLR2018}
E.~Grant, C.~Finn, S.~Levine, T.~Darrell, and T.~L. Griffiths, ``Recasting
  gradient-based meta-learning as hierarchical bayes,'' in \emph{ICLR}, 2018.

\bibitem{FranceschiICML18}
L.~Franceschi, P.~Frasconi, S.~Salzo, R.~Grazzi, and M.~Pontil, ``Bilevel
  programming for hyperparameter optimization and meta-learning,'' in
  \emph{ICML}, 2018.

\bibitem{LeeICML18}
Y.~Lee and S.~Choi, ``Gradient-based meta-learning with learned layerwise
  metric and subspace,'' in \emph{ICML}, 2018.

\bibitem{ZhangNIPS2018MetaGAN}
R.~Zhang, T.~Che, Z.~Grahahramani, Y.~Bengio, and Y.~Song, ``Metagan: An
  adversarial approach to few-shot learning,'' in \emph{NeurIPS}, 2018.

\bibitem{AntoniouICLR19}
A.~Antoniou, H.~Edwards, and A.~Storkey, ``How to train your maml,'' in
  \emph{ICLR}, 2019.

\bibitem{RusuICLR2019}
A.~A. Rusu, D.~Rao, J.~Sygnowski, O.~Vinyals, R.~Pascanu, S.~Osindero, and
  R.~Hadsell, ``Meta-learning with latent embedding optimization,'' in
  \emph{ICLR}, 2019.

\bibitem{VinyalsBLKW16}
O.~Vinyals, C.~Blundell, T.~Lillicrap, K.~Kavukcuoglu, and D.~Wierstra,
  ``Matching networks for one shot learning,'' in \emph{NIPS}, 2016.

\bibitem{MunkhdalaiICML2017}
T.~Munkhdalai and H.~Yu, ``Meta networks,'' in \emph{ICML}, 2017.

\bibitem{LiICML2018}
Z.~Li, F.~Zhou, F.~Chen, and H.~Li, ``Meta-sgd: Learning to learn quickly for
  few shot learning,'' in \emph{ICML}, 2018.

\bibitem{LopezPazNIPS17}
D.~Lopez{-}Paz and M.~Ranzato, ``Gradient episodic memory for continual
  learning,'' in \emph{NIPS}, 2017.

\bibitem{McCloskey1989}
M.~McCloskey and N.~J. Cohen, ``Catastrophic interference in connectionist
  networks: The sequential learning problem,'' in \emph{Psychology of learning
  and motivation}, 1989, pp. 3--17.

\bibitem{SnellSZ17}
J.~Snell, K.~Swersky, and R.~S. Zemel, ``Prototypical networks for few-shot
  learning,'' in \emph{NIPS}, 2017.

\bibitem{SungCVPR2018}
F.~Sung, Y.~Yang, L.~Zhang, T.~Xiang, P.~H.~S. Torr, and T.~M. Hospedales,
  ``Learning to compare: Relation network for few-shot learning,'' in
  \emph{CVPR}, 2018.

\bibitem{chen19closerfewshot}
W.-Y. Chen, Y.-C. Liu, Z.~Kira, Y.-C. Wang, and J.-B. Huang, ``A closer look at
  few-shot classification,'' in \emph{ICLR}, 2019.

\bibitem{BengioLCW09}
Y.~Bengio, J.~Louradour, R.~Collobert, and J.~Weston, ``Curriculum learning,''
  in \emph{ICML}, 2009.

\bibitem{ShrivastavaGG16}
A.~Shrivastava, A.~Gupta, and R.~B. Girshick, ``Training region-based object
  detectors with online hard example mining,'' in \emph{CVPR}, 2016.

\bibitem{OreshkinNIPS18}
B.~N. Oreshkin, P.~Rodr{\'{\i}}guez, and A.~Lacoste, ``{TADAM:} task dependent
  adaptive metric for improved few-shot learning,'' in \emph{NeurIPS}, 2018.

\bibitem{Hinton1987}
H.~E. Geoffrey and P.~C. David, ``Using fast weights to deblur old memories,''
  in \emph{CogSci}, 1987.

\bibitem{BertinettoICLR2019ridge}
L.~Bertinetto, J.~F. Henriques, P.~H.~S. Torr, and A.~Vedaldi, ``Meta-learning
  with differentiable closed-form solvers,'' in \emph{ICLR}, 2019.

\bibitem{SatorrasICLR2018graph}
V.~G. Satorras and J.~B. Estrach, ``Few-shot learning with graph neural
  networks,'' in \emph{ICLR}, 2018.

\bibitem{LiuICLR2019transductive}
Y.~Liu, J.~Lee, M.~Park, S.~Kim, and Y.~Yang, ``Learning to propagate labels:
  Transductive propagation network for few-shot learning,'' in \emph{ICLR},
  2019.

\bibitem{LiICML2019best1shotResult}
H.~Li, W.~Dong, X.~Mei, C.~Ma, F.~Huang, and B.~Hu, ``Lgm-net: Learning to
  generate matching networks for few-shot learning,'' in \emph{ICML}, 2019, pp.
  3825--3834.

\bibitem{LiCVPR2019bestResult}
H.~Li, D.~Eigen, S.~Dodge, M.~Zeiler, and X.~Wang, ``Finding task-relevant
  features for few-shot learning by category traversal,'' in \emph{CVPR}, 2019.

\bibitem{SantoroBBWL16}
A.~Santoro, S.~Bartunov, M.~Botvinick, D.~Wierstra, and T.~P. Lillicrap,
  ``Meta-learning with memory-augmented neural networks,'' in \emph{ICML},
  2016.

\bibitem{MishraICLR2018}
N.~Mishra, M.~Rohaninejad, X.~Chen, and P.~Abbeel, ``Snail: A simple neural
  attentive meta-learner,'' in \emph{ICLR}, 2018.

\bibitem{RaviICLR2017}
S.~Ravi and H.~Larochelle, ``Optimization as a model for few-shot learning,''
  in \emph{ICLR}, 2017.

\bibitem{LeeCVPR19svm}
K.~Lee, S.~Maji, A.~Ravichandran, and S.~Soatto, ``Meta-learning with
  differentiable convex optimization,'' in \emph{CVPR}, 2019.

\bibitem{YangICDM07}
J.~Yang, R.~Yan, and A.~G. Hauptmann, ``Adapting {SVM} classifiers to data with
  shifted distributions,'' in \emph{ICDM Workshops}, 2007.

\bibitem{WeiICML2018}
Y.~Wei, Y.~Zhang, J.~Huang, and Q.~Yang, ``Transfer learning via learning to
  transfer,'' in \emph{ICML}, 2018.

\bibitem{AmirCVPR18}
A.~R. Zamir, A.~Sax, W.~B. Shen, L.~J. Guibas, J.~Malik, and S.~Savarese,
  ``Taskonomy: Disentangling task transfer learning,'' in \emph{CVPR}, 2018.

\bibitem{Sun_2017_CVPR}
Q.~Sun, B.~Schiele, and M.~Fritz, ``A domain based approach to social relation
  recognition,'' in \emph{CVPR}, 2017.

\bibitem{Erhan10}
D.~Erhan, Y.~Bengio, A.~C. Courville, P.~Manzagol, P.~Vincent, and S.~Bengio,
  ``Why does unsupervised pre-training help deep learning?'' \emph{Journal of
  Machine Learning Research}, vol.~11, pp. 625--660, 2010.

\bibitem{HuangCVPR017}
J.~Huang, V.~Rathod, C.~Sun, M.~Zhu, A.~Korattikara, A.~Fathi, I.~Fischer,
  Z.~Wojna, Y.~Song, S.~Guadarrama, and K.~Murphy, ``Speed/accuracy trade-offs
  for modern convolutional object detectors,'' in \emph{CVPR}, 2017.

\bibitem{He_MaskRCNN17}
K.~He, G.~Gkioxari, P.~Doll{\'{a}}r, and R.~B. Girshick, ``Mask {R-CNN},'' in
  \emph{ICCV}, 2017.

\bibitem{ChenPAMI18}
L.~Chen, G.~Papandreou, I.~Kokkinos, K.~Murphy, and A.~L. Yuille, ``Deeplab:
  Semantic image segmentation with deep convolutional nets, atrous convolution,
  and fully connected crfs,'' \emph{{IEEE} Trans. Pattern Anal. Mach. Intell.},
  vol.~40, no.~4, pp. 834--848, 2018.

\bibitem{RohrbachNIPS13transfer}
M.~Rohrbach, S.~Ebert, and B.~Schiele, ``Transfer learning in a transductive
  setting,'' in \emph{NIPS}, 2013, pp. 46--54.

\bibitem{Keshari18}
R.~Keshari, M.~Vatsa, R.~Singh, and A.~Noore, ``Learning structure and strength
  of {CNN} filters for small sample size training,'' in \emph{CVPR}, 2018.

\bibitem{QiaoCVPR2018}
S.~Qiao, C.~Liu, W.~Shen, and A.~L. Yuille, ``Few-shot image recognition by
  predicting parameters from activations,'' in \emph{CVPR}, 2018.

\bibitem{ScottNIPS2018}
T.~R. Scott, K.~Ridgeway, and M.~C. Mozer, ``Adapted deep embeddings: {A}
  synthesis of methods for k-shot inductive transfer learning,'' in
  \emph{NeurIPS}, 2018.

\bibitem{SarafianosGNK17}
N.~Sarafianos, T.~Giannakopoulos, C.~Nikou, and I.~A. Kakadiaris, ``Curriculum
  learning for multi-task classification of visual attributes,'' in \emph{ICCV
  Workshops}, 2017.

\bibitem{WeinshallCA18}
D.~Weinshall, G.~Cohen, and D.~Amir, ``Curriculum learning by transfer
  learning: Theory and experiments with deep networks,'' in \emph{ICML}, 2018.

\bibitem{GravesICML2017}
A.~Graves, M.~G. Bellemare, J.~Menick, R.~Munos, and K.~Kavukcuoglu,
  ``Automated curriculum learning for neural networks,'' in \emph{ICML}, 2017.

\bibitem{KumarPK10}
M.~P. Kumar, B.~Packer, and D.~Koller, ``Self-paced learning for latent
  variable models,'' in \emph{NIPS}, 2010, pp. 1189--1197.

\bibitem{PentinaCVPR15}
A.~Pentina, V.~Sharmanska, and C.~H. Lampert, ``Curriculum learning of multiple
  tasks,'' in \emph{CVPR}, 2015.

\bibitem{JiangICML2018mentor}
L.~Jiang, Z.~Zhou, T.~Leung, L.~Li, and L.~Fei{-}Fei, ``Mentornet: Learning
  data-driven curriculum for very deep neural networks on corrupted labels,''
  in \emph{ICML}, 2018, pp. 2309--2318.

\bibitem{CanevetF16}
O.~Can{\'{e}}vet and F.~Fleuret, ``Large scale hard sample mining with monte
  carlo tree search,'' in \emph{CVPR}, 2016.

\bibitem{HarwoodGCRD17}
B.~Harwood, V.~Kumar, G.~Carneiro, I.~Reid, and T.~Drummond, ``Smart mining for
  deep metric learning,'' in \emph{ICCV}, 2017.

\bibitem{DalalT05}
N.~Dalal and B.~Triggs, ``Histograms of oriented gradients for human
  detection,'' in \emph{CVPR}, 2005.

\bibitem{Russakovsky2015}
O.~Russakovsky, J.~Deng, H.~Su, J.~Krause, S.~Satheesh, S.~Ma, Z.~Huang,
  A.~Karpathy, A.~Khosla, M.~Bernstein, A.~C. Berg, and L.~Fei-Fei, ``{ImageNet
  Large Scale Visual Recognition Challenge},'' \emph{International Journal of
  Computer Vision}, vol. 115, no.~3, pp. 211--252, 2015.

\bibitem{GidarisCVPR2018}
S.~Gidaris and N.~Komodakis, ``Dynamic few-shot visual learning without
  forgetting,'' in \emph{CVPR}, 2018.

\bibitem{MunkhdalaiICML18}
T.~Munkhdalai, X.~Yuan, S.~Mehri, and A.~Trischler, ``Rapid adaptation with
  conditionally shifted neurons,'' in \emph{ICML}, 2018.

\bibitem{CIFAR100}
A.~Krizhevsky, ``Learning multiple layers of features from tiny images,''
  \emph{University of Toronto}, 2009.

\bibitem{YeArXiv2018}
H.~Ye, H.~Hu, D.~Zhan, and F.~Sha, ``Learning embedding adaptation for few-shot
  learning,'' \emph{arXiv}, vol. 1812.03664, 2018.

\bibitem{IoffeICML15}
S.~Ioffe and C.~Szegedy, ``Batch normalization: Accelerating deep network
  training by reducing internal covariate shift,'' in \emph{ICML}, 2015.

\bibitem{ye2018learning}
H.-J. Ye, H.~Hu, D.-C. Zhan, and F.~Sha, ``Learning embedding adaptation for
  few-shot learning,'' \emph{arXiv preprint arXiv:1812.03664}, 2018.

\bibitem{kingma2014adam}
D.~P. Kingma and J.~Ba, ``Adam: A method for stochastic optimization,''
  \emph{arXiv}, vol. 1412.6980, 2014.

\end{thebibliography}
\bibliographystyle{IEEEtran}

%%%%%%%%%%%%%%%%%%%%%%%%%%%%%%%%%5
\begin{IEEEbiography}[{\includegraphics[width=1in,height=1.25in,clip,keepaspectratio]{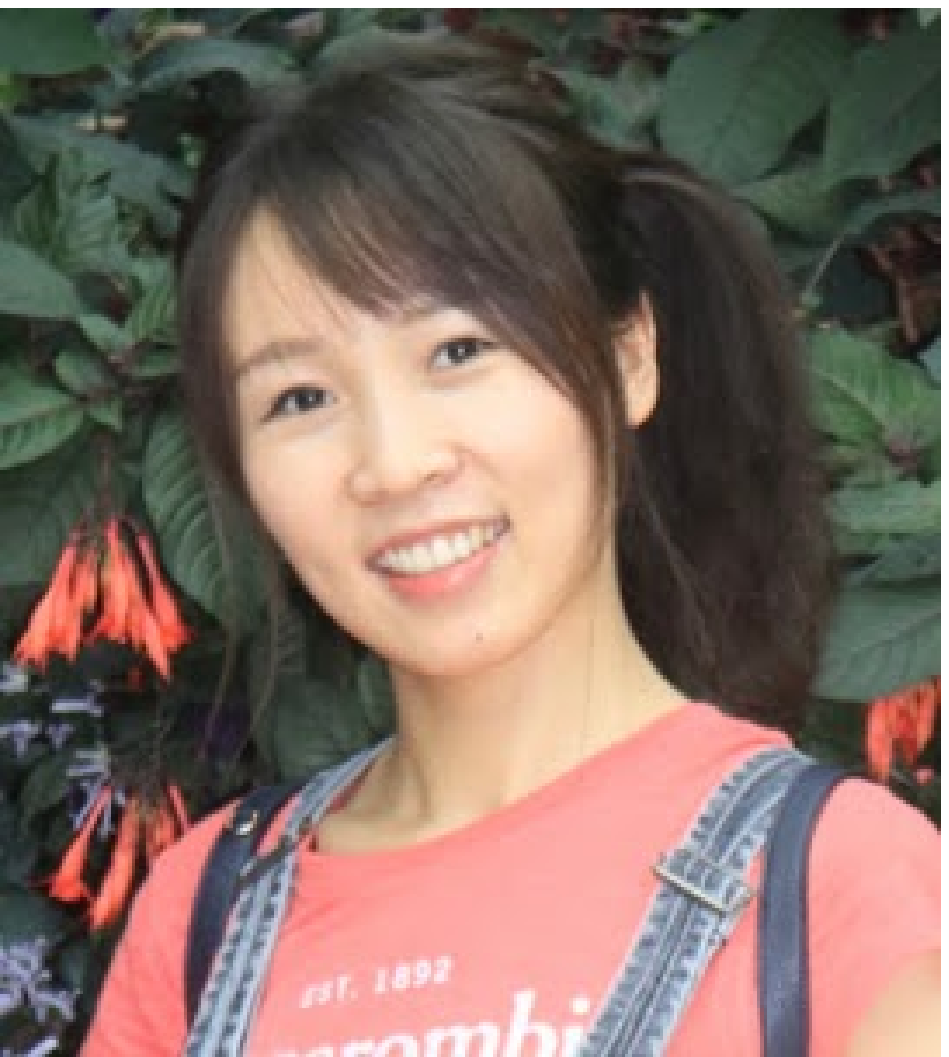}}]{Qianru Sun}
is an Assistant Professor in the School of Information Systems, Singapore Management University (since Aug 2019). Before that, she was a research fellow working with Prof. Tat-Seng Chua and leads the computer vision group in the NExT++ Lab at the National University of Singapore (since Apr 2018). Before that, she held the Lise Meitner Award Fellowship and was a post-doctoral researcher working with Prof.Dr. Bernt Schiele and Dr. Mario Fritz at the Max-Planck Institute for Informatics (MPII) for two years. She now has a visiting researcher position at MPII. In Jan 2016, she obtained her Ph.D. degree from Peking University, and her thesis was advised by Prof. Hong Liu. From Sep 2014 to Jan 2015, she was a visiting Ph.D. student advised by Prof. Tatsuya Harada at the University of Tokyo. Her research interests are computer vision and machine learning. Specific topics include image recognition, conditional image generation, meta-learning, and transfer learning.
\end{IEEEbiography}

\begin{IEEEbiography}[{\includegraphics[width=1in,height=1.25in,clip,keepaspectratio]{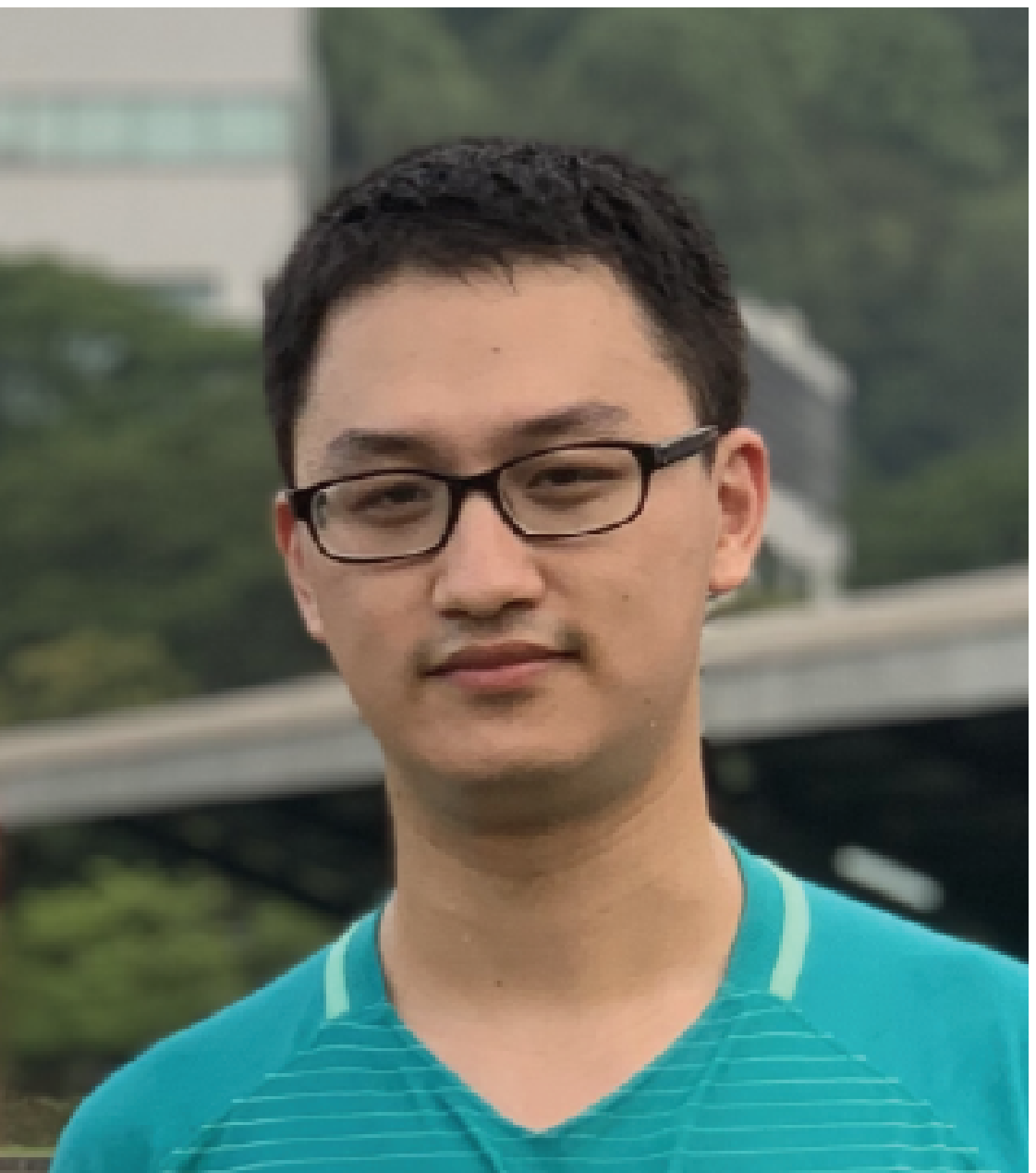}}]{Yaoyao Liu}
is a Ph.D. student at the School of Electrical and Information Engineering, Tianjin University advised by Prof. Yuting Su and Prof. An-An Liu. From June 2018, he is a visiting Ph.D. student at the School of Computing, National University of Singapore advised by Prof. Tat-Seng Chua and Prof. Qianru Sun. Before this, He obtained his bachelor degree at Qiushi Honors College, Tianjin University. 
His research lies in few-shot learning, meta learning and conditional image generation.
\end{IEEEbiography}

\begin{IEEEbiography}[{\includegraphics[width=1in,height=1.25in,clip,keepaspectratio]{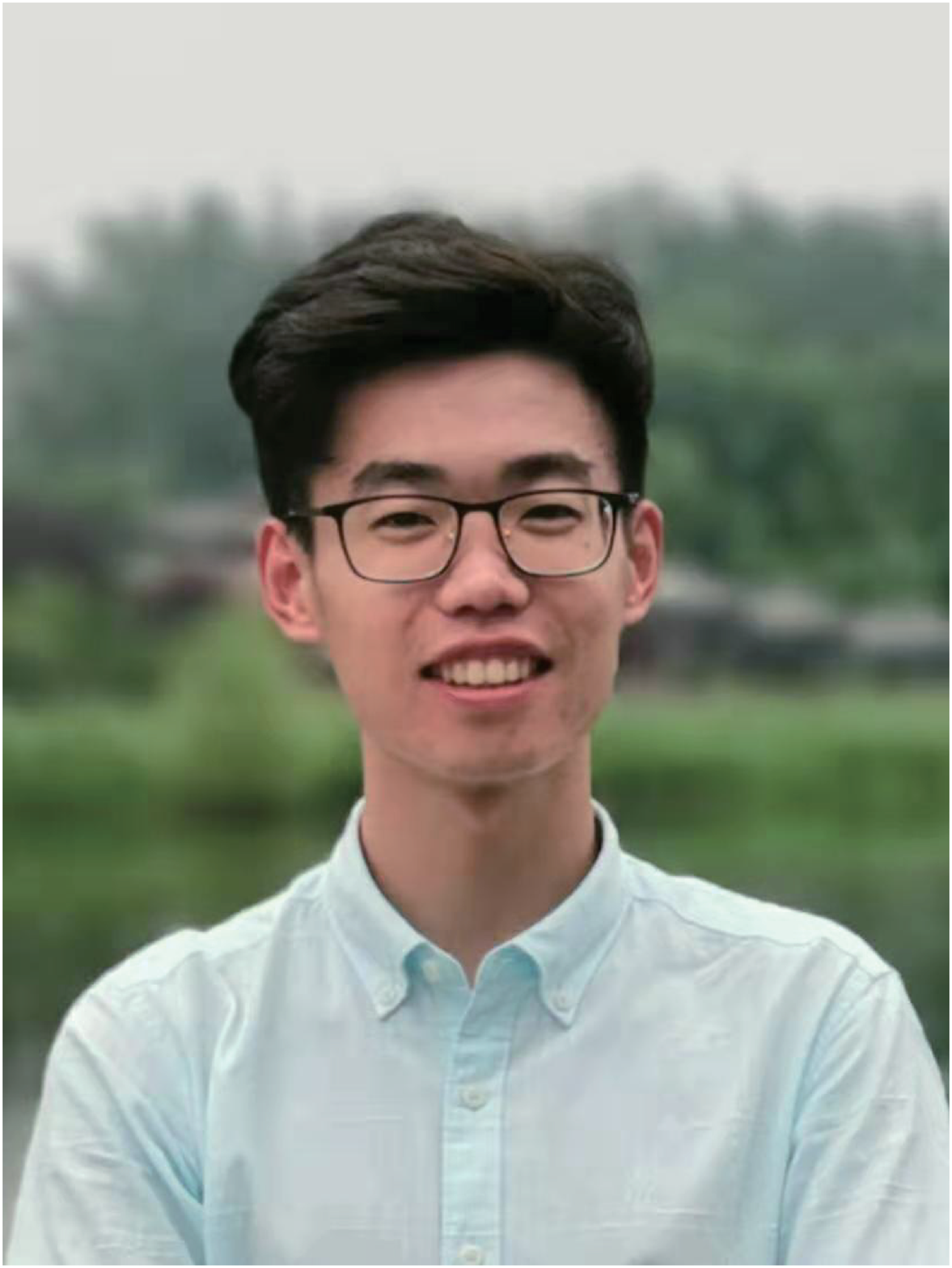}}]{Zhaozheng Chen}
is currently an intern and going to be a PhD student (from Jan 2020) at the School of Information Systems, Singapore Management University, advised by Prof. Qianru Sun. Before this, he obtained his bachelor degree at the School of Computer Science and Technology, Shandong University. His research interests are computer vision and machine learning.
\end{IEEEbiography}

\begin{IEEEbiography}[{\includegraphics[width=1in,height=1.25in,clip,keepaspectratio]{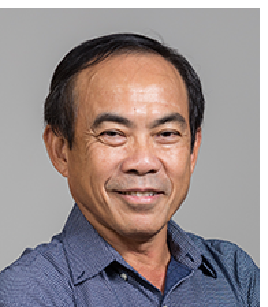}}]{Tat-Seng Chua}
is the KITHCT Chair Professor at the School of Computing, National University of Singapore. He is also the distinguish Visiting Professor of Tsinghua University. He was the Founding Dean of the School from 1998-2000. He is now the Director of a joint research Center between NUS and Tsinghua (NExT) to research into big unstructured multi-source multimodal data analytics. He holds a PhD degree from the University of Leeds, UK, since Feb 1983. Before that, he got his bachelor degree in the Civil Engineering and Computer Science, University of Leeds, UK, in Jun 1979. His main research interests are in multimedia information retrieval and social media analytics. In particular, his research focuses on the extraction, retrieval and question-answering of text, video and live media arising from the Web and social networks. 

\end{IEEEbiography}

\begin{IEEEbiography}[{\includegraphics[width=1in,height=1.25in,clip,keepaspectratio]{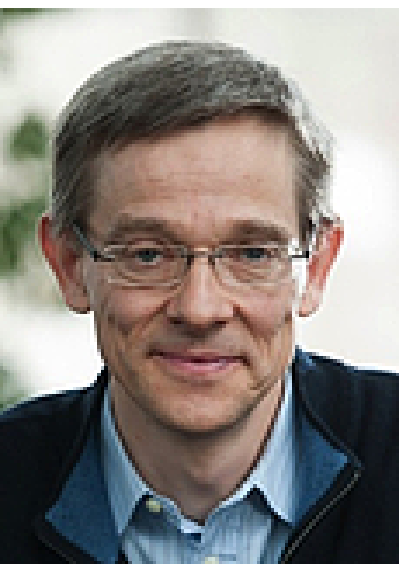}}]{Bernt Schiele}
has been Max Planck Director at MPI for Informatics and Professor at Saarland University since 2010.
He studied computer science at the University of Karlsruhe, Germany. He worked on his master thesis in the field of robotics in Grenoble, France, where he also obtained the "diplome d'etudes approfondies d'informatique". In 1994 he worked in the field of multi-modal human-computer interfaces at Carnegie Mellon University, Pittsburgh, PA, USA in the group of Alex Waibel. In 1997 he obtained his PhD from INP Grenoble, France under the supervision of Prof. James L. Crowley in the field of computer vision. The title of his thesis was "Object Recognition using Multidimensional Receptive Field Histograms". Between 1997 and 2000 he was postdoctoral associate and Visiting Assistant Professor with the group of Prof. Alex Pentland at the Media Laboratory of the Massachusetts Institute of Technology, Cambridge, MA, USA. From 1999 until 2004 he was Assistant Professor at the Swiss Federal Institute of Technology in Zurich (ETH Zurich). Between 2004 and 2010 he was Full Professor at the computer science department of TU Darmstadt.
\end{IEEEbiography}


% % insert where needed to balance the two columns on the last page with
% %\newpage



%\vfill


\end{document}